\documentclass{article}

\usepackage{arxiv}

\usepackage[utf8]{inputenc} % allow utf-8 input
\usepackage[T1]{fontenc}    % use 8-bit T1 fonts
\usepackage{hyperref}       % hyperlinks
\usepackage{url}            % simple URL typesetting
\usepackage{booktabs}       % professional-quality tables
\usepackage{amsfonts}       % blackboard math symbols
\usepackage{nicefrac}       % compact symbols for 1/2, etc.
\usepackage{microtype}      % microtypography
\usepackage[title]{appendix}
\usepackage{graphicx}
\usepackage[bf]{caption}

\usepackage{doi}
\usepackage{etoolbox}
\usepackage[natbibapa]{apacite}
\AtBeginDocument{}
\renewcommand{\APACrefnote}[1]{}

\newtoggle{bibdoi}
\newtoggle{biburl}
\makeatletter
\newsavebox{\bib@url}
\newsavebox{\bib@doi}

\undef{\APACrefURL}
\undef{\endAPACrefURL}
\undef{\APACrefDOI}
\undef{\endAPACrefDOI}

\newenvironment{APACrefURL}
  {\global\toggletrue{biburl}\lrbox\bib@url}
  {\endlrbox}

\newenvironment{APACrefDOI}
  {\global\toggletrue{bibdoi}\lrbox\bib@doi}
  {\endlrbox}

\newcommand{\printinfo}{
  \iftoggle{bibdoi}{\usebox{\bib@doi}}{\usebox{\bib@url}}
  \togglefalse{bibdoi}
}

\AtBeginEnvironment{thebibliography}{
\pretocmd{\PrintBackRefs}{%
  \iftoggle{bibdoi}
    {\iftoggle{biburl}{\unskip\unskip}{}\usebox{\bib@doi}}
    {\iftoggle{biburl}{Retrieved from \usebox{\bib@url}}}{}
  \togglefalse{bibdoi}\togglefalse{biburl}%
}{}{}}
\makeatother
\title{ELEV-VISION: Automated Lowest Floor Elevation Estimation from Segmenting Street View Images }
\date{} 					% Or removing it

\renewcommand{\headeright}{}
\renewcommand{\undertitle}{}
\renewcommand{\shorttitle}{}

\hypersetup{
pdftitle={ELEV-VISION: Automated Lowest Floor Elevation Estimation from Segmenting Street View Images},
pdfsubject={},
}

\begin{document}
\maketitle

\begin{center}
% authors go here:
{\Large
Yu-Hsuan Ho\textsuperscript{a,*},
Cheng-Chun Lee\textsuperscript{a},
Nicholas D. Diaz\textsuperscript{b},
Samuel D. Brody\textsuperscript{b},
Ali Mostafavi\textsuperscript{a}
\par}

\bigskip
\textsuperscript{a} Urban Resilience.AI Lab, Zachry Department of Civil and Environmental Engineering,\\ Texas A\&M University, College Station, TX\\
\textsuperscript{b} Department of Marine and Coastal Environmental Science,\\ Texas A\&M University at Galveston, Galveston, TX\\
\vspace{6pt}
\textsuperscript{*} corresponding author, email: yuhsuanho@tamu.edu
\\
\end{center}
\bigskip
\begin{abstract}
We propose an automated lowest floor elevation (LFE) estimation algorithm based on computer vision techniques to leverage the latent information in street view images. Flood depth-damage models use a combination of LFE and flood depth for determining flood risk and extent of damage to properties. We used image segmentation for detecting door bottoms and roadside edges from Google Street View images. The characteristic of equirectangular projection with constant spacing representation of horizontal and vertical angles allows extraction of the pitch angle from the camera to the door bottom. The depth from the camera to the door bottom was obtained from the depthmap paired with the Google Street View image. LFEs were calculated from the pitch angle and the depth. The testbed for application of the proposed method is Meyerland (Harris County, Texas). The results show that the proposed method achieved mean absolute error of 0.190 m (1.18 \%) in estimating LFE. The height difference between the street and the lowest floor (HDSL) was estimated to provide information for flood damage estimation. The proposed automatic LFE estimation algorithm using Street View images and image segmentation provides a rapid and cost-effective method for LFE estimation compared with the surveys using total station theodolite and unmanned aerial systems. By obtaining more accurate and up-to-date LFE data using the proposed method, city planners, emergency planners and insurance companies could make a more precise estimation of flood damage.
\end{abstract}

% keywords can be removed

% =======note=======
% \citep{}: with ()
% \citet{}: author name only
% \ref{fig:fig1}: figure
    % \begin{figure}
    % 	\centering
    %     \includegraphics[width=0.85\linewidth]{fig1.png}
    %     \caption{Schematic .}
    % 	\label{fig:fig1}
    % \end{figure}
% \ref{sec:Results}: section
    % \section{Results}
    % \label{sec:Results}
% section hierarchy
    % \section{}
    % \subsection{}
    % \subsubsection{}
% =======note=======

\section{Introduction}
\label{sec:1}
Floods are among the most costly natural disasters, especially in coastal areas \citep{aerts_evaluating_2014, guo_large-scale_2022, michel-kerjan_we_2015, stromberg_natural_2007}. In the United States, floods accounted for 40\% of natural disasters between 1900 and 2015 \citep{cigler_us_2017}. Besides weather and terrain, flood damage is also influenced by human development. Urban built environment has been identified to exacerbate flood damage \citep{brody_identifying_2008}. The cost of flood damage can be enormous, with more than \$72 billion USD in claims filed by National Flood Insurance Program (NFIP) policyholders between 1990 and 2022 \citep{fima_fima_nodate}. The ability to estimate flood damages is critical for mitigation and response actions \citep{aerts_pathways_2018}. One key factor for estimating flood damage is the lowest floor elevation (LFE) \citep{bodoque_flood_2016, taghinezhad_imputation_2020, xian_optimal_2017, zarekarizi_neglecting_2020}. Flood depth-damage models use a combination of LFE and flood depth for determining flood risk and projecting extent of damages to properties. The absence of reliable and up-to-date LFE data is a major roadblock in proper assessment of flood risk to properties and rapid assessment of flood damages.

%%To moderate flood losses, the Federal Emergency Management Agency (FEMA) recommends elevating houses in flood zones in delineated flood zones to reduce flood losses. \citep{fema_coastal_2000}. FEMA specifies a  base flood elevation representing the 100-year flood level, with a 1-percent probability of being equaled or exceeded by the flood level in any given year. To avoid severe flood losses, FEMA recommends elevating houses at least 1 foot above the BFE, known as a 1-foot freeboard \citep{fema_appendix_2020}. Knowing the LFE of buildings is one factor to be considered in determining the level of protection needed to mitigate flood damage to residences. Inundation depth, another component in flood-damage assessment  \citep{de_moel_effect_2011, taghinezhad_imputation_2020}. Inundation depth is the extent of flooding experienced by a building \citep{de_ruig_micro-scale_2020, jamali_rapid_2018}. This study estimates both the LFE and the height difference between the street and the lowest floor (HDSL), providing critical information for homeowners and agencies to make informed decisions and mitigate potential flood damage.

%%The NFIP mandates that the lowest floor of all new structures and substantially improved buildings in flood-prone areas must be elevated to or above the BFE specified by FEMA \citep{horn_introduction_2018}. 

The current methods of estimating and measuring  are costly, time consuming, and resource intensive. One approach is direct measurement by survey contractors using a total station theodolite. The resulting elevation certificate must be completed, sealed, and submitted to the jurisdiction's building code department by a licensed surveyor, engineer, or architect \citep{fema_appendix_2020}. This type of  manual inspection, however, is costly and labor-intensive. Previous literature has shown the use of drones and street view images to estimate LFE. For example, \citet{diaz_deriving_2022} used drones and photogrammetric methods to derive LFE in coastal areas; drone aviation, however, is subject to regulatory and training limitations. Mapping LFE at a larger scale (such as an entire city) using drones could be costly due to the aforementioned limitations. LiDAR (light detection and ranging) data has been used for similar measurement in previous literature \citep{FENG2022101759, BONCZAK2019126, MEESUK2017239}. \citet{FENG2022101759} used LiDAR data captured by vehicles for detecting building openings, such as doors and windows, for mapping building flood risk. Capturing LiDAR data, however, is more expensive than capturing images. Another approach is the use of street view images (SVI), such as Google Street View, are publicly available. \citet{ning_exploring_2022} used Google Street View images and a computer vision algorithm, YOLO-v5, to estimate LFE. \citet{gao_exploring_2023} analyzed flood mitigation governance based on \citet{ning_exploring_2022}'s LFE estimation method; however, these studies lacked estimation of the height difference between the street and the lowest floor (HDSL), which is critical information for flood vulnerability estimation. In addition, the previous studies used bounding boxes, rectangles to frame objects in images, to mark door shapes. Bounding boxes, however, are not able to discern precise object shapes in some situations, such as variation in viewpoints and distortion in projection, since shapes are perceived only as rectangles. When the viewpoint of an image is not perfectly aligned with the target door, the door shape will be a parallelogram instead of a rectangle. When a panorama is presented on a plane, inherent distortion causes door edges to appear as non-straight. To address the limitation of bounding boxes to delineate door shapes from the images with viewpoint variation and projection distortion, the previous method reprojects street view equirectangular panoramas to perspective images before object detection. Image reprojection, however, results in loss of information. Furthermore, bounding boxes can provide only straight lines, yet streets are not always straight, making this method unsuitable for HDSL estimation. To address these limitations, we propose an automated LFE and HDSL estimation algorithm using computer-vision techniques directly on street view panoramas without image reprojection and demonstrate the application of the proposed algorithm using data from Meyerland in Harris County, Texas (USA).
%due to the limitations related to object-detection bounding boxes in providing precise object shapes, the previous studies may have failed to maintain the robustness of LFE estimation with viewpoint variation and camera distortion.

\begin{figure}
    \centering
    \includegraphics[width=1\textwidth]{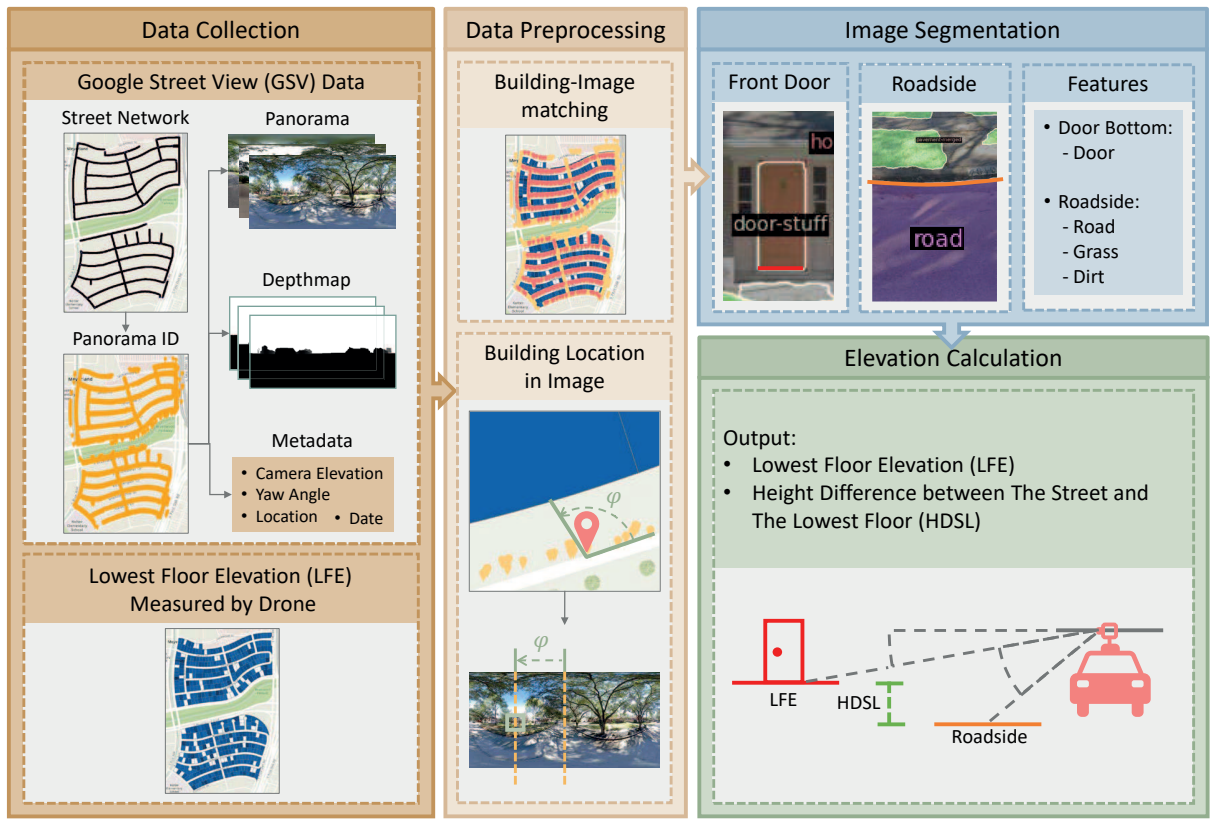}
    \caption{Overview of the Elev-Vision method to estimate LFE and HDSL. The elevation estimation method comprises four steps: data collection, data preprocessing, image segmentation, and elevation calculation. Data collection: the street map should be downloaded first to determine the required street view images on the streets of interest; street view panoramas, depthmaps, and metadata are collected based on the collected panorama IDs. Data preprocessing: street view images are matched with houses based on addresses; the image with the most visible front door is selected for each house; the house of interest is automatically located in the image based on the bearing angle from the camera to the house. Image segmentation: the front door bottom  and the road elevation are detected.  If the road is not detected, the grass and the dirt are used as the secondary feature for roadside extraction. Elevation calculation: LFE and HDSL are calculated by the depth and the pitch angle.}
    \label{fig:flowchart}
\end{figure}

%%The objective of this study is to leverage street view images, such as those available through Google Street View, along with computer vision techniques to efficiently estimate (1) the lowest floor elevation (LFE) and (2) the height difference between the street and the lowest floor (HDSL). 

 Figure \ref{fig:flowchart} depicts an overview of the proposed method. The first step involves collecting images and depthmaps from Google Street View, which are then preprocessed to align with each building in the study area. Image segmentation techniques are  then utilized to detect doors, which can be used to determine LFEs. Similarly, objects such as roads, grass, and dirt are used to determine the elevation of the roadside, providing information necessary to calculate HDSL. The \hyperref[sec:2]{data and methods} section describes the data sources and the methodology used in this research. The \hyperref[sec:3]{results and discussion} section presents the estimation of the two measurements when implementing computer-vision techniques on street view images. It discusses the corresponding error then provides an in-depth analysis of the developed estimation algorithm. The \hyperref[sec:4]{closing remarks} section summarizes the contributions, limitations, and suggestions for future works of this study.

\section{Data and Methods}
\label{sec:2}
\subsection{Study Area and Ground Truth Data}
\begin{figure}[ht]
    \centering
    \includegraphics[width=0.7\textwidth]{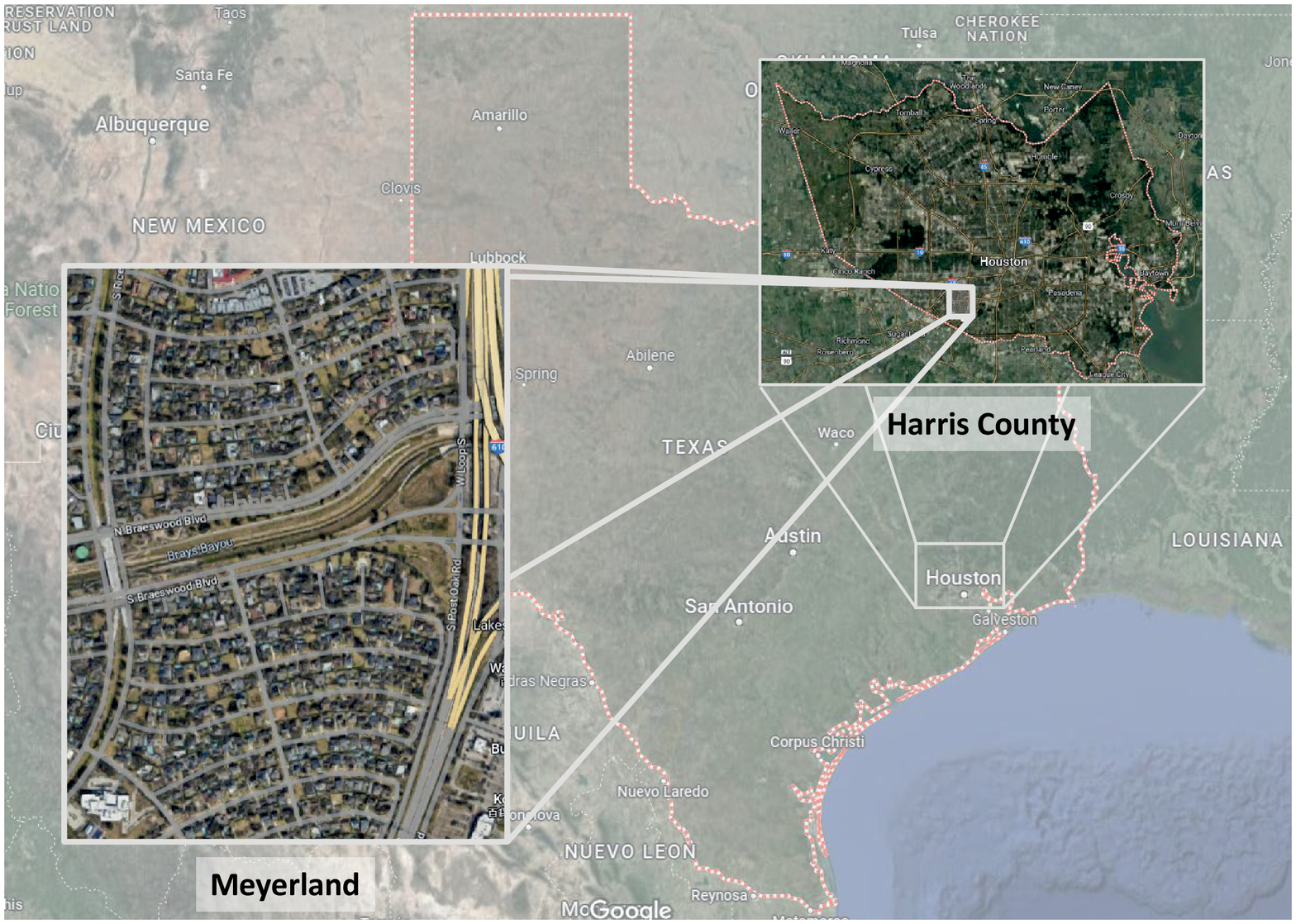}
    \caption{Meyerland, a flood-prone community in southern Harris County, Texas, was the testbed for this research.}
    \label{fig:study_area}
\end{figure}

\begin{figure}[ht]
    \centering
    \includegraphics[width=0.7\textwidth]{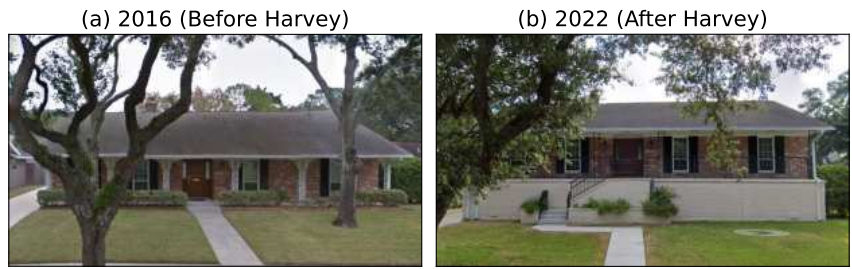}
    \caption{Example of a house elevated after Hurricane Harvey. (a) The house in 2016 (Before Hurricane Harvey). (b) The house in 2022 (After Hurricane Harvey).}
    \label{fig:elevated_examples}
\end{figure}

% Introduce the baseline data
In this study, we used the LFE measured by an unmanned aerial system-based photogrammetry as ground truth to evaluate the LFE derived by our proposed street view image-based method. The drone-based LFE measurements achieved 0.16 m (6.3 in) mean absolute error (MAE) to elevation certificates (ECs) from the Federal Emergency Management Agency (FEMA). Detailed description of drone measurement is in \citet{diaz_deriving_2022}.The drone-based LFE was measured using the bottom of front, side, and back doors. On the other hand, the LFE determined in ECs is the lowest enclosed area, including garages and basements \citep{fema_appendix_2020}. Since the drone-based LFE is the available data whose definition of LFE is the closest to our definition, the bottom of front doors, the drone-based LFE is adopted as ground truth in this study. The inconsistency between the LFE in ECs and the LFE derived by image-based methods has been indicated in several studies \citep{diaz_deriving_2022, ning_exploring_2022}.
% Introduce the study area
The study area is in Meyerland, a flood-prone community in Harris County, Texas. About \$0.68 billion USD in claims was filed by National Flood Insurance Program (NFIP) policyholders in Meyerland between 1990 and 2022 \citep{fima_fima_nodate}. In recent years, lots of buildings in this community were reconstructed or elevated to avoid being flooded. An example of a house in Meyerland elevated after Hurricane Harvey is shown in Figure \ref{fig:elevated_examples}. The flood vulnerability of Meyerland, however, is deteriorating with the expansion of the floodplain extent \citep{juan_comparing_2020}. The need for flood risk evaluation and the LFE difference among new and old buildings make this community a suitable testbed for evaluating the Elev-Vision method. Figure \ref{fig:study_area} shows the study area.

\subsection{Street View Images}
% Introduce the source of data and how you select the image for each building
Street view imagery, an accessible urban data source, makes available easily accessible high-quality images. \citet{PERRETT2023101968} developed a roadside verge biodiversity classification model using street view images for roadside verge surveys. \citet{LIU2023101924} used street view images and a street tree classification model to establish street tree inventories. \citet{ALIZADEH2022101730} combined flood gauge data and flooded street view images for flood depth estimation. \citet{LIU2023104096} used street view images for three-dimensional streetscape-feature estimation to explore the relationship between three-dimensional streetscapes and vehicle use. \citet{BILJECKI2021104217} provides a comprehensive review on the applications of street view images in studies related to spatial and urban analysis. 
 
 The road network of Meyerland was first downloaded from OpenStreetMap \citep{OpenStreetMap} for street view image collection. Google Street View images, with available related depthmaps and metadata, offer stable quality across the desired area and time period. Google Street View panoramas and their metadata were collected based on streetview \citep{letchford_streetview_2023}, a Python module for downloading Google Street View images. The depthmaps were decoded based on GSVPanoDepth \citep{hausler_gsvpanodepth_nodate}. The resolutions of most panoramas are 8192 x 16384 pixels and 6656 x 13312 pixels. The resolution of depthmaps is 256 x 512 pixels. Collected metadata includes image location, address, camera elevation, camera yaw angle, and image captured date. Camera elevations are in WGS84 (World Geodetic System 1984) meters above mean sea level (MSL), the same as drone-based LFEs.
 
 The research area contains 483 houses. Due to house reconstruction during LFE measurement time, ground truth data for only 424 LFE (87.78\%) were available for the analysis. Among the houses with LFE ground truth, 409 (96.46\%) of houses' street view images were available. Among these street view images, 197 (48.17\%) houses' front doors were visible. Some of the images, however, were not matched with the LFE ground truth because of reconstruction. Specifically, some street view images showed newly constructed houses, whereas the LFE ground truth was for the houses before reconstruction. Despite the availability of older images, only the images taken after 2015 were used in this study with consideration of image resolution. After removal of unmatched images, 185 (93.91\%) images were available. The numbers of houses passing the data filter in each step are listed in Table \ref{tab:data}.

\begingroup

\setlength{\tabcolsep}{10pt} % Default value: 6pt
\renewcommand{\arraystretch}{1.5} % Default value: 1

\begin{table}
\caption{Number of houses passing the filter in each step.}
\centering
\begin{tabular}{ccc}
\cline{1-3}
Step & Data Filter                                      & Number of Houses \\
\cline{1-3}
1    & Houses with LFE ground truth                     & 424              \\
2    & Houses with an SVI                               & 409              \\
3    & Houses with a visible front door in the SVI      & 197              \\
4    & Houses with matched LFE ground truth and the SVI & 185              \\
5    & Houses with the detected door bottom in the SVI\textsuperscript{*} & 136              \\
\cline{1-3}
\multicolumn{3}{l}{\textsuperscript{*} Detailed explanation of the number of houses in Step 5 is provided in Section \ref{sec:2.3}}.
\end{tabular}
\label{tab:data}
\end{table}
\endgroup

%\subsection{Measured Lowest Floor Elevation}
%\label{sec:2.3}

\subsection{Image Segmentation}
\label{sec:2.3}
% Lit review on image segmentation and the reason of chosing the current method
The computer vision-based method is used extensively in civil infrastructure inspection \citep{lee_assessing_2022, zhang_concrete_2020}, building damage assessment \citep{kaur_large-scale_2023, gao_deep_2018, sajedi_vibration-based_2020}, and other related applications. In this study, we used image segmentation, a computer vision technique that involves pixel classification, to detect building objects that could be used for finding lowest floor elevation. Image segmentation comprises three categories: semantic segmentation, instance segmentation, and panoptic segmentation \citep{9356353}. Semantic segmentation is categorical labeling for each pixel to create a categorical mask for each category. Instance segmentation creates an individual mask for each object, which can be view as the combination of object detection and semantic segmentation. Panoptic segmentation is a combination of the semantic and image segmentation; for object categories, instance segmentation is conducted; for background categories, semantic segmentation is conducted.

Image segmentation has been applied to street view images in city environmental research applications, such as built environment \citep{JEON2023104631, HE2023104189, MA2021103086, chen_automatic_2023, CHIANG2023104679, MIDDEL2019122, KI2021103920}, acoustic environment \citep{ZHAO2023101915}, air pollution \citep{qi_using_2021}, and route planning \citep{GUAN2023101975, li_pedestrian_2019}. \citet{JEON2023104631} assessed walking environments of public housing using street view images and image segmentation; \citet{qi_using_2021} applied image segmentation to street view images with a combination of land-use regression models for air pollution prediction; \citet{chen_automatic_2023} adopted image segmentation and street view images to assess public open spaces; \citet{CHIANG2023104679} applied image segmentation to street view images for estimating sky view factor and green view index; \citet{GUAN2023101975} analyzed the complexity involved in decision-making at road intersections by the application of street view images and image segmentation; \citet{HE2023104189} employed image segmentation on street view images for extracting human perceptions to built environment and evaluating the potential in urban renewal; \citet{ZHAO2023101915} combined image processing, object detection, image segmentation, and scene recognition to explore urban soundscapes. Image segmentation has also been utilized in other urban and structural studies \citep{KIDO2021101281, WU2019100936, PAN2022101767, ZOU2021101421,CZERNIAWSKI2020101131, HOSSEINI2023101950}. \citet{PAN2022101767} used image segmentation on structure facades for identifying thermal anomalies. \citet{KIDO2021101281} applied image segmentation and enhanced mixed reality to estimating designed landscapes to assess environmental impact of construction projects. \citet{WU2019100936} used image segmentation and point cloud to extract road pothole data.

Different image segmentation models have been developed in recent years \citep{long_fully_2015, chen_rethinking_2017}. In this study, we adopted OneFormer \citep{Jain_2023_CVPR} , a transformer-based multi-task image segmentation framework, to detect door bottoms and roadsides. OneFormer was selected because of its good segmentation performance on COCO-stuff dataset \citep{caesar_coco-stuff_2018}. The OneFormer model pretrained by COCO-stuff dataset is utilized. We also experimented with other datasets for door segmentation, but the COCO-stuff dataset was the only dataset on which front doors were able to be segmented. Most datasets could only be used for indoor door segmentation.

Through COCO-stuff pretrained OneFormer, 136 (73.51\%) of these visible door bottoms in the aforementioned 185 street view images were correctly segmented while 100\% of roadsides were segmented.

\subsection{LFE Estimation}
%\label{sec:2.5}
% Introduce the algorithm to calculate LFE. Please include as many equations as possible.
\subsubsection{House Localization in Panorama}
The house of interest needs to be located from among the many houses in the panorama. First, the bearing angle from the camera to the house \(\beta_{house} (radian)\) is calculated using Eq. \ref{eq:bear}, 

\begin{equation}\label{eq:bear}
\beta_{house}  = atan2(X, Y)
\end{equation}

\begin{equation}
X = sin(lon_{house} - lon_c)\cdot cos(lat_{house})
\end{equation}

\begin{equation}
Y = cos(lat_c)\cdot sin(lat_{house}) - sin(lat_c)\cdot cos(lat_{house})\cdot cos(lon_{house} - lon_c)
\end{equation}

in which \((lat_c, lon_c)\) is the camera location and \((lat_{house}, lon_{house})\) is the house location. Since Google street view panoramas are equirectangular projection images, for the property upon which vertical and horizontal angles are projected, constant spacing lines can be utilized to transform between angles and pixels. The horizontal pixel \(p_x\) of the house is calculated by Eq. \ref{eq:loc},

\begin{equation}\label{eq:loc}
p_x = \left\{
\begin{array}{rcl}
\frac{W_{img}}{2} + \frac{\Delta\beta}{180} \cdot \frac{W_{img}}{2} & & {\beta_{yaw} \leq \beta_{house} \leq \beta_{yaw} + 180}\\\\
\frac{W_{img}}{2} - \frac{\Delta\beta}{180} \cdot \frac{W_{img}}{2} & & {else}
\end{array} \right.
\end{equation}

in which \(\beta_{yaw} (degree)\) is the yaw angle of the camera, \(\beta_{house} (degree)\) is the bearing angle from the camera to the house, \(\Delta\beta\) is the angle between the yaw direction the direction from the camera to the house, and \(W_{img}\) is the width of the panorama.

From house location, the possible location range of the front door can be determined.  We set the possible location range of the front door for image segmentation as \(\pm\) 45 degrees from the bearing angle of the house,  \(\left[\beta_{house}-45^\circ, \beta_{house}+45^\circ\right]\). If the front door is not detected, the possible location range is horizontally translated \(\pm\) 22.5 degrees. House localization not only prevents house mismatching but also accelerates the process time of image segmentation.
 
\subsubsection{Door-bottom Extraction}
Next, we conduct image segmentation to get the category matrix. Eq. \ref{eq:door} extracts pixels of the door bottom from the door mask. The pixel

\begin{equation}\label{eq:door}
(p_x, p_y) \in P_{db} \quad\forall\quad \left[p_x \in P_{x, door} \quad and \quad (p_y = \max p'_y \quad\forall\quad p'_y \in P_{y, door}(p_x))\right]
\end{equation}

in which \(P_{db}\) is the set of the door bottom pixels, \(P_{x, door}\) is the set of the door's horizontal pixels, and \(P_{y, door}(p_x)\) is the set of the door's vertical pixels given the horizontal pixel.

\subsubsection{Elevation Calculation}
To obtain LFE, we first calculate the height difference between the camera and the door bottom \(\Delta h_{db, c}\) with the pitch angle from the camera to the door bottom \(\Delta\theta_{db, c}\) and the depth from the camera to the door bottom \(d_{db, c}\) by Eq. \ref{eq:height}. 

\begin{equation}\label{eq:height}
\Delta h_{db, c} = d_{db, c} \cdot sin(\Delta\theta_{db, c})
\end{equation}

\begin{figure}[ht]
    \centering
    \includegraphics[width=1\textwidth]{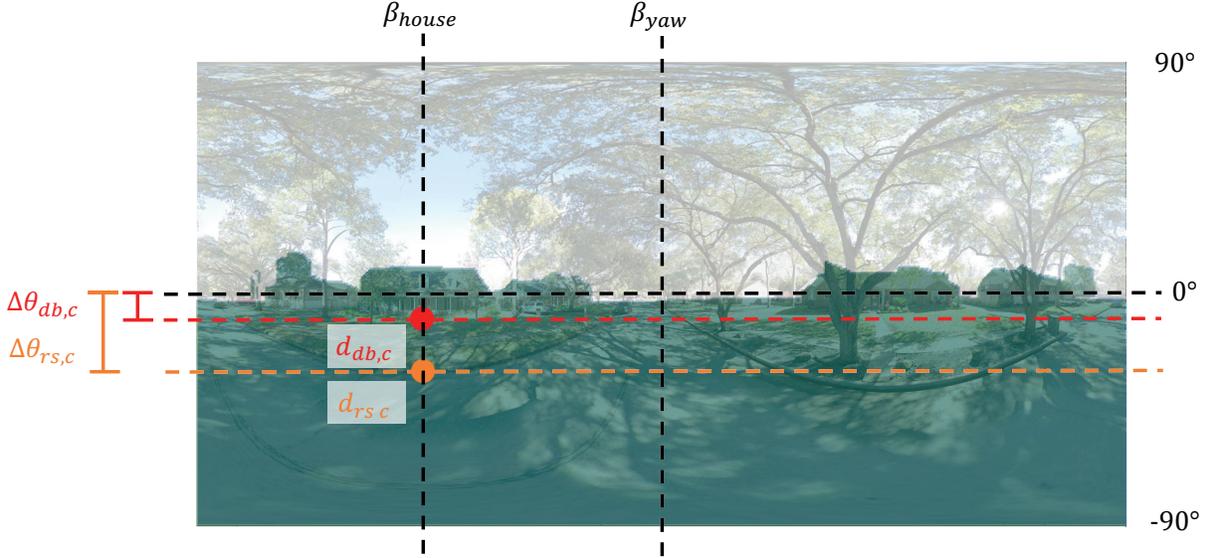}
    \caption{The example of the depthmap with the spherical coordinate system demonstration. Pixels with darker masks represent shorter depths. \(\beta_{yaw} (degree)\) is the yaw angle of the camera, \(\beta_{house} (degree)\) is the bearing angle from the camera to the house, \(\Delta\theta_{db, c}\) is the pitch angle from the camera to the door bottom, \(\Delta\theta_{rs, c}\) is the pitch angle from the camera to the roadside, \(d_{db, c}\) is the depth from the camera to the door bottom,  \(d_{rs, c}\) is the depth from the camera to the roadside. Google Street View panoramas are equirectangular projection images, of which vertical and horizontal angles are projected to constant spacing lines. This characteristic makes it simple to transform between angles and pixels.}
    \label{fig:depthmap}
\end{figure}

The depth from the camera to the door bottom \(d_{db, c}\) is extracted from the depthmap based on the door bottom pixels obtained in the previous step and the pitch angle from the camera to the door bottom \(\Delta\theta_{db, c}\) is calculated by Eq. \ref{eq:pitch},

\begin{equation}\label{eq:pitch}
\Delta\theta_{db, c} = (\frac{H_{img}}{2} - p_y) \cdot \frac{180}{H_{img}}
\end{equation}

in which \(H_{img}\) is the height of the panorama and \(p_y\) is the vertical pixel of the door bottom. The depth and the pitch angle are demonstrated in Figure \ref{fig:depthmap}.

Then, the estimated LFE is performed using
\begin{equation}\label{eq:LFE}
LFE = CE + \Delta h_{db, c}
\end{equation}

in which \(CE\) represents camera elevation.

To provide more robust estimation, we use the median of the set of the door bottom elevations \(\{E_{db}\}\) derived from the set of the door bottom pixels \(P_{db}\) as the estimated LFE instead of using only one point \((p_x, p_y)\). However, due to the view angle of images, the door in the image is a quadrilateral rather than a perfect square. LFE will be overestimated if we simply use the median of elevations derived from the set of the door bottom pixels \(P_{db}\) as the estimated LFE. To solve this problem, the outliers of door bottom elevations are removed from the set of the door bottom elevations \(\{E_{db}\}\) first. Then, as shown in \ref{fig:LFE_dist}, the detected front doors are labeled as completely detected or partially detected. Specifically, we labeled the detected door bottoms for all houses at the subset \(\{E'_{db}\}\) containing only the 25\%, 50\%, or 75\% lowest door bottom elevations. For completely detected front doors, \(\{E'_{db}\} = \{E_{db}\}\). Last, we further removed the elevations higher than the median of the subset of door bottom elevations \(\{E'_{db}\}\) to construct another subset \(\{E''_{db}\}\)  then calculated the median of \(\{E'_{db}\}\) as the estimated LFE.

\subsection{Height Difference between the Street and the Lowest Floor Estimation}
% Introduce the algorithm to calculate HDSL. Include all equations.

To estimate HDSL, we first calculated the elevation of the roadside \(RE\) near the house with the pitch angle from the camera to the roadside \(\Delta\theta_{rs, c}\) and the depth from the camera to the roadside \(d_{rs, c}\). The calculation of \(RE\) is similar to the calculation of LFE. Roads, grass, and dirt are the features used to detect roadsides.. Grass, and dirt are secondary features for roadside detection because roads are not always correctly segmented. When using secondary features to detect roadsides, we selected pixels which were 20 pixels vertically lower than the edge of the feature mask to represent the roadside more accurately. Then, HDSL was calculated using Eq. \ref{eq:HDSL}.

\begin{equation}\label{eq:HDSL}
HDSL = LFE - RE
\end{equation}

\section{Results and Discussion}
\label{sec:3}
\begin{figure}%[ht]
    \centering    \includegraphics[width=1\textwidth]{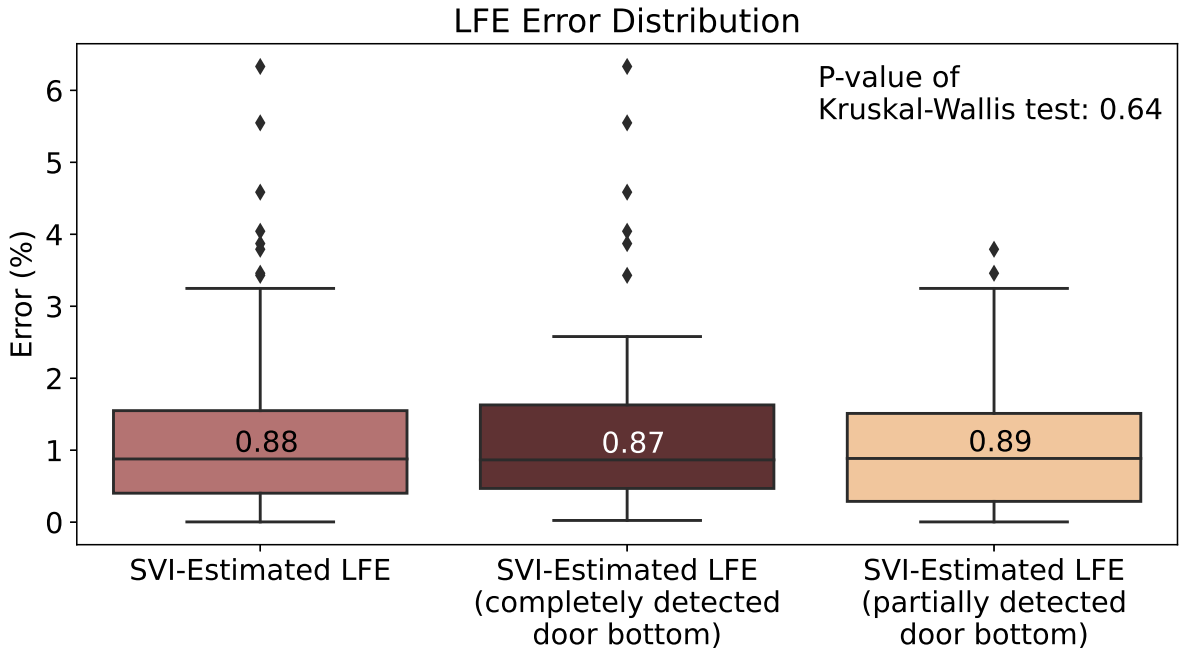}
    \caption{Lowest floor elevation error distribution. The error range of LFE of the completely detected door bottoms is narrower than that of the partially detected door bottoms, despite more outliers in completely detected door bottoms. The medians of error of the three subsets of SVI-estimated LFE show no significant difference. The p-value of a Kruskal-Wallis test among them is 0.64, greater than the significance level of 0.1. The result suggests that partial occlusion of the door bottom has no significant influence on our proposed LFE estimation algorithm. Regardless of whether if the door bottom is completely or partially visible, the algorithm performs well.}
    \label{fig:LFE_dist}
\end{figure}

% FFE estimation results and 'First Floor Height from street' (we need a name for it) results
SVI-estimated LFSs were compared to drone-measured LFE as the ground truth. The error distribution is shown in Figure \ref{fig:LFE_dist}. As the front doors of houses are not always completely visible, the data were classified as door bottom completely visible or partially visible to evaluate the influence of image segmentation performance on LFE estimation. Figure \ref{fig:LFE_dist} demonstrates that the LFE error range is narrower for completely detected door bottoms than for partially detected ones, even though there are more outliers among the former. However, there is no significant difference in terms of their medians, as indicated by a Kruskal-Wallis test with a p-value of 0.64, which is greater than the significance level of 0.1. In other words, the LFE estimation pipeline developed in this study shows similar performance whether the door bottom is fully visible or not. The mean absolute error of SVI-estimated LFE is 0.190 m (1.18\%).

\begin{figure}%[ht]
    \centering\includegraphics[width=0.7\textwidth]{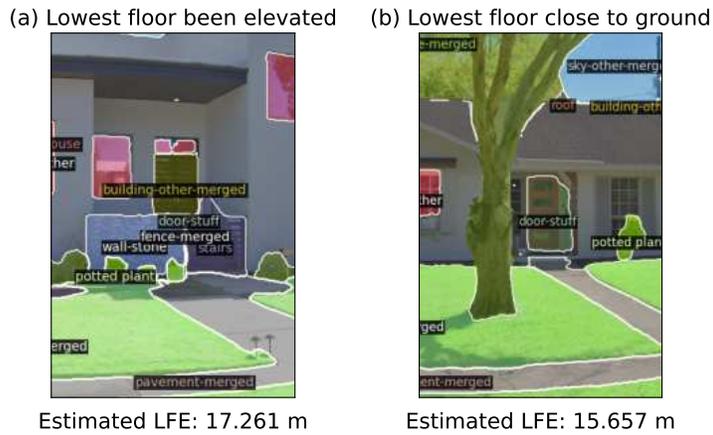}
    \caption{The examples of estimated LFEs for two adjacent houses. (a) The house with a staircase in front of its front door. (b) The house with a front door closer to the ground. The lowest floor of (a) are significantly elevated, resulting in a higher LFE. Conversely, the lowest floor of (b) is closer to the ground level, resulting in a lower LFE. The distinction between the two can be effectively captured by the proposed SVI method.}
    \label{fig:LFE_examples}
\end{figure}

To demonstrate the effectiveness of the Elev-Vision method, we provide examples of two adjacent houses with significantly different LFEs on Imogene Street, a flat street within the study area, in Figure \ref{fig:LFE_examples}, along with the corresponding LFE estimates obtained via SVI. Thus, the difference in their LFEs is due mainly to the height of their lowest floor. Specifically, the house in Figure \ref{fig:LFE_examples}a has a staircase in front of its front door, indicating that the lowest floor is significantly elevated, while the house in Figure \ref{fig:LFE_examples}b has a front door closer to the ground, indicating a lower LFE. The proposed method using SVI allows observation of this difference and effectively estimate the LFE for each building.

\begin{figure}%[ht]
    \centering
    \includegraphics[width=0.5\textwidth]{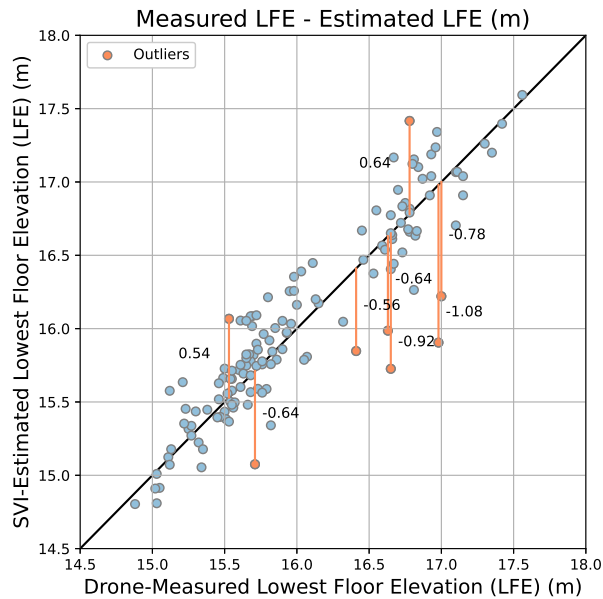}
    \caption{Scatter plot of SVI-estimated LFE by drone-measured LFE. Red scatters are the outliers, of which the error is greater than the upper quartile \((Q_3)\) plus 1.5 times the interquartile range (IQR). The red line shows the difference between SVI-estimated LFE and drone-measured LFE of the outliers. Most of the outliers  underestimate LFE compared to drone-measured value, which is conservative for the application to evaluate the risk of flood damage estimation because a lower LFE value is more likely to be evaluated as at the risk of flood damage.}
    \label{fig:LFE_fit}
\end{figure}

Figure \ref{fig:LFE_fit} compares the SVI-estimated LFE with the drone-measured LFE, revealing that most outliers are underestimated cases. Specifically, when there are large errors, the SVI estimations tends to provide lower LFE values than the drone measurement. Since a lower LFE value is more likely to be evaluated as at-risk of flood damage, it is a more conservative result in the context of flood damage estimation. A possible reason of outliers is low resolution of depthmaps. The resolution of the depthmap is only 256 x 512 pixels, which may not be sufficient to distinguish the depth difference for objects at longer distance. It reveals that the resolution of depthmaps is critical for elevation estimation.

\begin{figure}%[ht]
    \centering
    \includegraphics[width=0.5\textwidth]{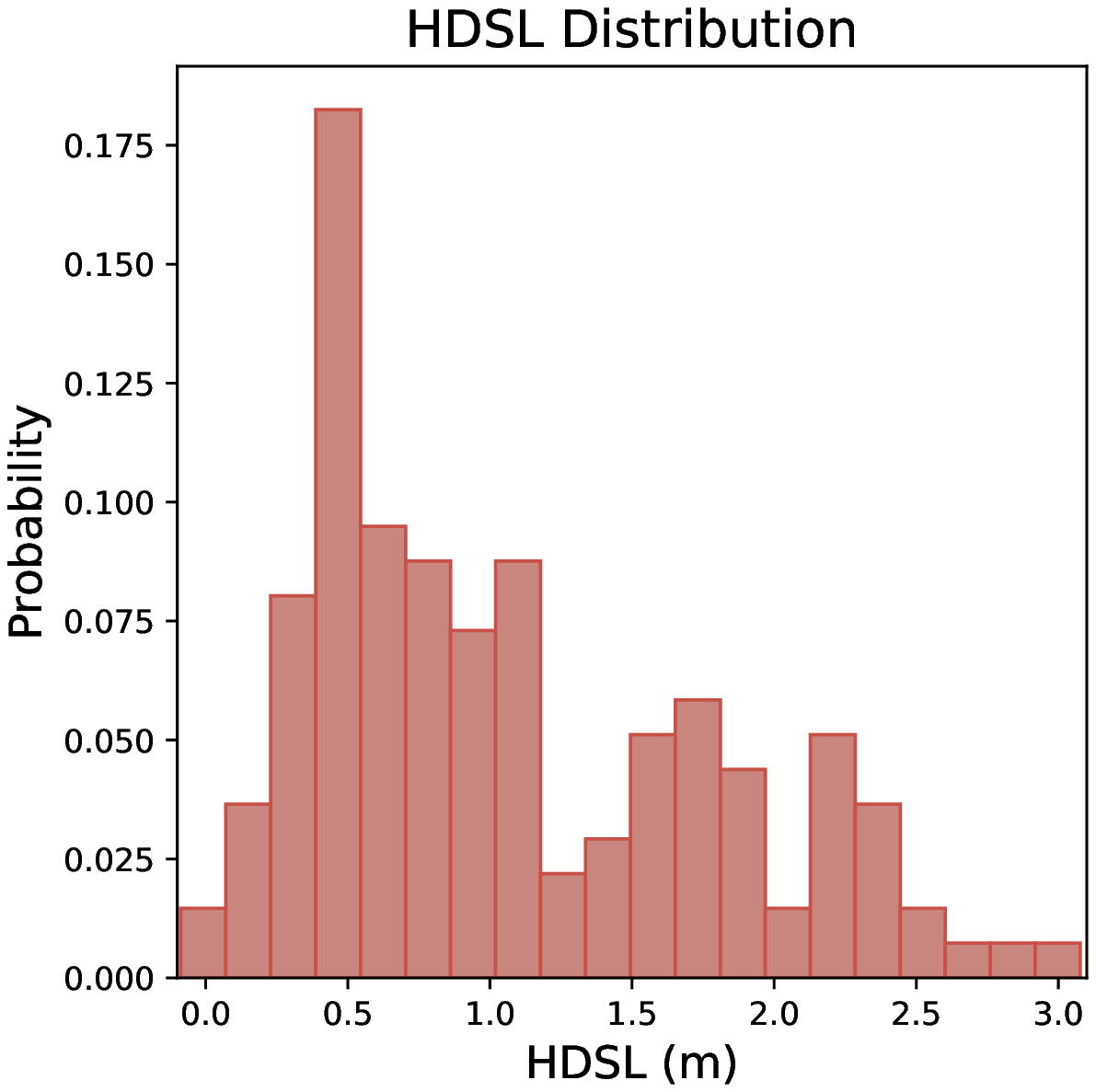}
    \caption{Distribution of height difference between the street and the lowest floor (HDSL). The average HDSL in this research area is 1.074 m (3.524 ft) and the median of HDSL is 0.874 m (2.867 ft). 70\% of the buildings in the study area have an HDSL greater than 0.536 m (1.759 ft). 7.30\% of houses have an HDSL lower than 1 ft (0.305 m), indicating higher potential for damage during flood events.}
    \label{fig:HDSL_dist}
\end{figure}

In addition to LFE, this study estimates HDSL to provide information regarding the risk of flood damage when juxtaposed with inundation depth information. Figure \ref{fig:HDSL_dist} shows the distribution of the estimated HDSL. The average HDSL in this study area is 1.074 m (3.524 ft), while the median of HDSL is 0.874 m (2.867 ft). In the study area, 70\% of houses have HDSLs higher than 0.536 m  (1.759 ft), indicating that they may avoid severe flood damage if the flood inundation depth is less than 0.536 m (1.759 ft). However, 7.30\% of houses have an HDSL lower than 0.305 m (1 ft), indicating higher potential for damage during flood events. 

\begin{figure}%[ht]
    \centering
    \includegraphics[width=1\textwidth]{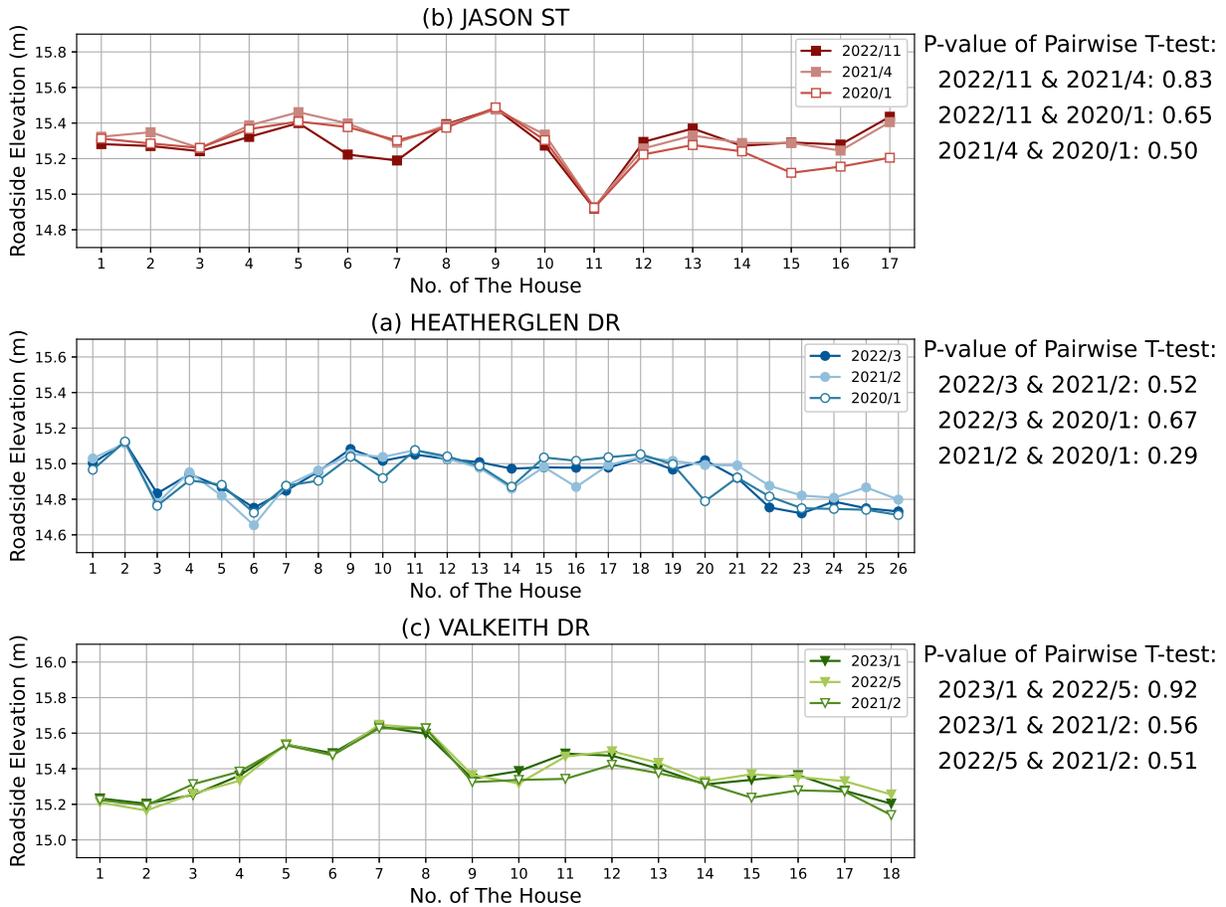}
    \caption{Estimated roadside elevations of three selected streets in three different time periods. Houses are numbered from east to west. The p-values of pairwise T-test for three streets are all greater than the significance level of 0.1, indicating no significant difference among the estimated roadside elevations in the same location across different time periods.}
    \label{fig:eroad_samestreet}
\end{figure}

\begin{figure}%[ht]
    \centering
    \includegraphics[width=0.7\textwidth]{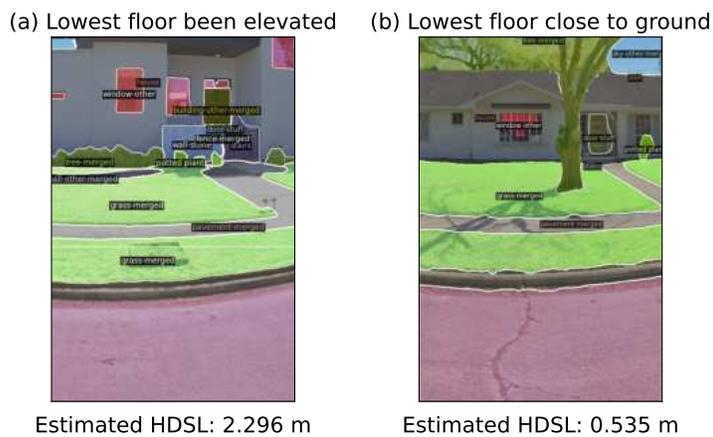}
    \caption{The examples of estimated HDSLs for two adjacent houses. (a) The house with a staircase at its front door. (b) The house with a front door closer to the ground. The LFE of house (a) is higher than the LFE of house (b), resulting in a higher HDSL, since their roadside elevations are similar. The distinction between the two can be effectively captured by the proposed SVI method.}
    \label{fig:HDSL_examples}
\end{figure}

Due to the unavailability of ground truth for roadside elevations, we selected three streets with  SVIs obtained from several time periods to validate the accuracy of our roadside elevation estimates. Figure \ref{fig:eroad_samestreet} shows the estimated roadside elevations for each street. We conducted a pairwise T-test to analyze the estimated roadside elevations from the images captured in different time periods. The p-values of pairwise T-test for three streets are all greater than the significance level of 0.1, indicating no significant difference among the estimated roadside elevations in the same location across different time periods. In addition, the same examples of the two adjacent buildings with significant differences in first floor heights from Figure \ref{fig:LFE_examples} illustrate SVI-estimated HDSLs (Figure \ref{fig:HDSL_examples}). These sample results demonstrate the effectiveness and robustness of the proposed method in estimating HDSL, providing critical information to residents and decision-makers regarding the flood vulnerability of homes.

\section{Closing Remarks}
\label{sec:4}
% Outliers, house having invisible doors, non-rectangular doors, etc.
% \subsection{Contribution}
 This paper presents the Elev-Vision method that integrates Google Street View images and computer vision techniques to estimate the lowest floor elevation (LFE) and the height difference between the street and the lowest floor. To the authors' best knowledge, this work is the first LFE estimation method based on the application of image segmentation, which can provide precise object shapes from street view images. The proposed method achieves the mean absolute error of 19 cm (7.5 in) for LFEs compared to drone-based ground truth, outperforming the previous method \citep{ning_exploring_2022} based on object detection and street view images, which had an error of 22 cm (8.7 in). The advantage of using image segmentation masks rather than object detection bounding boxes for door bottom detection is that image segmentation maintains good performance even with viewpoint variations and projection distortion. With the application of image segmentation, we were able to maximize the use of the rich information in street view panoramas without loss of information due to image reprojection, filling the gap in the previous LFE estimation method using object detection. In addition, our results suggest that partial occlusion of door bottom has no significant influence on the proposed LFE estimation method (p-value > 0.1). Compared to \citet{ning_exploring_2022}'s method treating the door height as the reference height with an assumed value to calculate the height difference between the camera and the door bottom, our proposed method calculates the height difference without any assumed value, reducing the errors from the assumption of reference height. Furthermore, with the advantage of image segmentation in providing masks for objects with any shape, roadsides are able to be identified and marked regardless of curvature and image distortions. Relying on the marked roadside, this study calculates another critical indicator, HDSL, to help mitigate severe flood damage, filling the gap in the extant literature in investigating HDSL and overcoming the limitation in the previous method based on object detection in estimating HDSL. With this indicator, property owners and emergency managers can develop appropriate plans based on the predicted flood inundation depths. The practical contribution of this work lies in providing an efficient and cost-effective approach for estimating lowest floor elevation. Traditional LFE surveys employing a total station theodolite are time-consuming, labor-intensive, and costly, while the current measuring technology adopting drones has limitations in regulatory and professional training and it requires manual derivation of LFEs within photogrammetry software for each structure, both of which present hindrances for real-time or large-scale measurement. Moreover, to the best of our knowledge, the height difference between the street and the lowest floor has not been investigated in the extant literature. Our study fills this research gap and offers critical information for the accuracy improvement of LFEs and other flood risk evaluation at the individual structure scale. 

It should be noted that there are two limitations in the proposed Elev-Vision method: the resolution of the depthmap and the visibility of the front door. Since the resolution of the depthmap is 256 x 512 pixels, while the vertical and horizontal fields of view are 180 degrees and 360 degrees, the depth difference within 0.7 degrees may not be captured, which is acceptable for close objects but problematic for distant objects. In other words, the applicable distance of the proposed method is limited. On the other hand, as we currently use the door bottom as the lowest floor, the proposed method can be applied only to the street view images in which the front door is visible. For houses whose front doors are hidden, features such as windows and stairs may be utilized as second features for LFE and HDSL estimation. The estimation algorithm, however, would become more complex since the uncertainty of these features' locations is much higher than that of the front door's location. 

This work can be furthered in some aspects. Due to the current limitation in the quality of street view depthmaps, depth estimation methods can be applied to street view images to obtain the depthmaps with the same size as the input street view images. In addition, exploring the feasibility of the other building features can be the direction of future works. The proposed Elev-Vision method has many applications on vertical measurements besides LFE estimation. For example, this method can be applied to structure height estimation. It is beneficial to built environment evaluation research and urban canyon effect research. It can also be applied to sidewalk height estimation for sidewalk accessibility assessment research.

\section{Data Availability}
The data that support the findings of this study are available from Google Street View Data.

\section{Code Availability}
The code that supports the findings of this study is available from the corresponding author upon request.

\section{Acknowledgements}
The authors would like to acknowledge funding support from  the National Science Foundation under CRISP 2.0 Type 2,  grant 1832662,  and the Texas A\&M X-Grant Presidential Excellence Fund. Any opinions, findings, conclusions, or recommendations expressed in this research are those of the authors and do not necessarily reflect the view of the funding agencies.

\bibliography{ref, ref_zotero}

\begin{thebibliography}{}

\bibitem [\protect \citeauthoryear {%
J\BPBI C.~Aerts%
\ \protect \BOthers {.}}{%
J\BPBI C.~Aerts%
\ \protect \BOthers {.}}{%
{\protect \APACyear {2018}}%
}]{%
aerts_pathways_2018}
\APACinsertmetastar {%
aerts_pathways_2018}%
\begin{APACrefauthors}%
Aerts, J\BPBI C.%
, Barnard, P\BPBI L.%
, Botzen, W.%
, Grifman, P.%
, Hart, J\BPBI F.%
, De~Moel, H.%
\BDBL {}Sadrpour, N.%
\end{APACrefauthors}%
\unskip\
\newblock
\APACrefYearMonthDay{2018}{}{}.
\newblock
{\BBOQ}\APACrefatitle {Pathways to resilience: adapting to sea level rise in
  {Los} {Angeles}} {Pathways to resilience: adapting to sea level rise in {Los}
  {Angeles}}.{\BBCQ}
\newblock
\APACjournalVolNumPages{Annals of the New York Academy of
  Sciences}{1427}{1}{1--90}.
\newblock
\begin{APACrefURL}
  [{2023-05-09}]\url{https://onlinelibrary.wiley.com/doi/abs/10.1111/nyas.13917}
  \end{APACrefURL}
\newblock
\begin{APACrefDOI} \doi{10.1111/nyas.13917} \end{APACrefDOI}
\PrintBackRefs{\CurrentBib}

\bibitem [\protect \citeauthoryear {%
J\BPBI C\BPBI J\BPBI H.~Aerts%
\ \protect \BOthers {.}}{%
J\BPBI C\BPBI J\BPBI H.~Aerts%
\ \protect \BOthers {.}}{%
{\protect \APACyear {2014}}%
}]{%
aerts_evaluating_2014}
\APACinsertmetastar {%
aerts_evaluating_2014}%
\begin{APACrefauthors}%
Aerts, J\BPBI C\BPBI J\BPBI H.%
, Botzen, W\BPBI J\BPBI W.%
, Emanuel, K.%
, Lin, N.%
, de Moel, H.%
\BCBL {}\ \BBA {} Michel-Kerjan, E\BPBI O.%
\end{APACrefauthors}%
\unskip\
\newblock
\APACrefYearMonthDay{2014}{{\APACmonth{05}}}{}.
\newblock
{\BBOQ}\APACrefatitle {Evaluating {Flood} {Resilience} {Strategies} for
  {Coastal} {Megacities}} {Evaluating {Flood} {Resilience} {Strategies} for
  {Coastal} {Megacities}}.{\BBCQ}
\newblock
\APACjournalVolNumPages{Science}{344}{6183}{473--475}.
\newblock
\begin{APACrefURL}
  [{2023-05-08}]\url{https://www.science.org/doi/full/10.1126/science.1248222}
  \end{APACrefURL}
\newblock
\begin{APACrefDOI} \doi{10.1126/science.1248222} \end{APACrefDOI}
\PrintBackRefs{\CurrentBib}

\bibitem [\protect \citeauthoryear {%
Alizadeh%
\ \protect \BOthers {.}}{%
Alizadeh%
\ \protect \BOthers {.}}{%
{\protect \APACyear {2022}}%
}]{%
ALIZADEH2022101730}
\APACinsertmetastar {%
ALIZADEH2022101730}%
\begin{APACrefauthors}%
Alizadeh, B.%
, Li, D.%
, Hillin, J.%
, Meyer, M\BPBI A.%
, Thompson, C\BPBI M.%
, Zhang, Z.%
\BCBL {}\ \BBA {} Behzadan, A\BPBI H.%
\end{APACrefauthors}%
\unskip\
\newblock
\APACrefYearMonthDay{2022}{}{}.
\newblock
{\BBOQ}\APACrefatitle {Human-centered flood mapping and intelligent routing
  through augmenting flood gauge data with crowdsourced street photos}
  {Human-centered flood mapping and intelligent routing through augmenting
  flood gauge data with crowdsourced street photos}.{\BBCQ}
\newblock
\APACjournalVolNumPages{Advanced Engineering Informatics}{54}{}{101730}.
\newblock
\begin{APACrefURL}
  \url{https://www.sciencedirect.com/science/article/pii/S1474034622001884}
  \end{APACrefURL}
\newblock
\begin{APACrefDOI} \doi{https://doi.org/10.1016/j.aei.2022.101730}
  \end{APACrefDOI}
\PrintBackRefs{\CurrentBib}

\bibitem [\protect \citeauthoryear {%
Biljecki%
\ \BBA {} Ito%
}{%
Biljecki%
\ \BBA {} Ito%
}{%
{\protect \APACyear {2021}}%
}]{%
BILJECKI2021104217}
\APACinsertmetastar {%
BILJECKI2021104217}%
\begin{APACrefauthors}%
Biljecki, F.%
\BCBT {}\ \BBA {} Ito, K.%
\end{APACrefauthors}%
\unskip\
\newblock
\APACrefYearMonthDay{2021}{}{}.
\newblock
{\BBOQ}\APACrefatitle {Street view imagery in urban analytics and GIS: A
  review} {Street view imagery in urban analytics and gis: A review}.{\BBCQ}
\newblock
\APACjournalVolNumPages{Landscape and Urban Planning}{215}{}{104217}.
\newblock
\begin{APACrefURL}
  \url{https://www.sciencedirect.com/science/article/pii/S0169204621001808}
  \end{APACrefURL}
\newblock
\begin{APACrefDOI} \doi{https://doi.org/10.1016/j.landurbplan.2021.104217}
  \end{APACrefDOI}
\PrintBackRefs{\CurrentBib}

\bibitem [\protect \citeauthoryear {%
Bodoque%
\ \protect \BOthers {.}}{%
Bodoque%
\ \protect \BOthers {.}}{%
{\protect \APACyear {2016}}%
}]{%
bodoque_flood_2016}
\APACinsertmetastar {%
bodoque_flood_2016}%
\begin{APACrefauthors}%
Bodoque, J\BPBI M.%
, Guardiola-Albert, C.%
, Aroca-Jimenez, E.%
, Eguibar, M\BPBI A.%
\BCBL {}\ \BBA {} Martínez-Chenoll, M\BPBI L.%
\end{APACrefauthors}%
\unskip\
\newblock
\APACrefYearMonthDay{2016}{{\APACmonth{07}}}{}.
\newblock
{\BBOQ}\APACrefatitle {Flood {Damage} {Analysis}: {First} {Floor} {Elevation}
  {Uncertainty} {Resulting} from {LiDAR}-{Derived} {Digital} {Surface}
  {Models}} {Flood {Damage} {Analysis}: {First} {Floor} {Elevation}
  {Uncertainty} {Resulting} from {LiDAR}-{Derived} {Digital} {Surface}
  {Models}}.{\BBCQ}
\newblock
\APACjournalVolNumPages{Remote Sensing}{8}{7}{604}.
\newblock
\begin{APACrefURL} [{2023-05-08}]\url{https://www.mdpi.com/2072-4292/8/7/604}
  \end{APACrefURL}
\newblock
\begin{APACrefDOI} \doi{10.3390/rs8070604} \end{APACrefDOI}
\PrintBackRefs{\CurrentBib}

\bibitem [\protect \citeauthoryear {%
Bonczak%
\ \BBA {} Kontokosta%
}{%
Bonczak%
\ \BBA {} Kontokosta%
}{%
{\protect \APACyear {2019}}%
}]{%
BONCZAK2019126}
\APACinsertmetastar {%
BONCZAK2019126}%
\begin{APACrefauthors}%
Bonczak, B.%
\BCBT {}\ \BBA {} Kontokosta, C\BPBI E.%
\end{APACrefauthors}%
\unskip\
\newblock
\APACrefYearMonthDay{2019}{}{}.
\newblock
{\BBOQ}\APACrefatitle {Large-scale parameterization of 3D building morphology
  in complex urban landscapes using aerial LiDAR and city administrative data}
  {Large-scale parameterization of 3d building morphology in complex urban
  landscapes using aerial lidar and city administrative data}.{\BBCQ}
\newblock
\APACjournalVolNumPages{Computers, Environment and Urban
  Systems}{73}{}{126-142}.
\newblock
\begin{APACrefURL}
  \url{https://www.sciencedirect.com/science/article/pii/S0198971518300176}
  \end{APACrefURL}
\newblock
\begin{APACrefDOI} \doi{https://doi.org/10.1016/j.compenvurbsys.2018.09.004}
  \end{APACrefDOI}
\PrintBackRefs{\CurrentBib}

\bibitem [\protect \citeauthoryear {%
Brody%
\ \protect \BOthers {.}}{%
Brody%
\ \protect \BOthers {.}}{%
{\protect \APACyear {2008}}%
}]{%
brody_identifying_2008}
\APACinsertmetastar {%
brody_identifying_2008}%
\begin{APACrefauthors}%
Brody, S\BPBI D.%
, Zahran, S.%
, Highfield, W\BPBI E.%
, Grover, H.%
\BCBL {}\ \BBA {} Vedlitz, A.%
\end{APACrefauthors}%
\unskip\
\newblock
\APACrefYearMonthDay{2008}{}{}.
\newblock
{\BBOQ}\APACrefatitle {Identifying the impact of the built environment on flood
  damage in Texas} {Identifying the impact of the built environment on flood
  damage in texas}.{\BBCQ}
\newblock
\APACjournalVolNumPages{Disasters}{32}{1}{1-18}.
\newblock
\begin{APACrefURL}
  \url{https://onlinelibrary.wiley.com/doi/abs/10.1111/j.1467-7717.2007.01024.x}
  \end{APACrefURL}
\newblock
\begin{APACrefDOI} \doi{https://doi.org/10.1111/j.1467-7717.2007.01024.x}
  \end{APACrefDOI}
\PrintBackRefs{\CurrentBib}

\bibitem [\protect \citeauthoryear {%
Caesar%
\ \protect \BOthers {.}}{%
Caesar%
\ \protect \BOthers {.}}{%
{\protect \APACyear {2018}}%
}]{%
caesar_coco-stuff_2018}
\APACinsertmetastar {%
caesar_coco-stuff_2018}%
\begin{APACrefauthors}%
Caesar, H.%
, Uijlings, J.%
\BCBL {}\ \BBA {} Ferrari, V.%
\end{APACrefauthors}%
\unskip\
\newblock
\APACrefYearMonthDay{2018}{{\APACmonth{03}}}{}.
\newblock
\APACrefbtitle {{COCO}-{Stuff}: {Thing} and {Stuff} {Classes} in {Context}.}
  {{COCO}-{Stuff}: {Thing} and {Stuff} {Classes} in {Context}.}
\newblock
\APACaddressPublisher{}{arXiv}.
\newblock
\begin{APACrefURL} [{2023-05-15}]\url{http://arxiv.org/abs/1612.03716}
  \end{APACrefURL}
\newblock
\APACrefnote{arXiv:1612.03716 [cs]}
\newblock
\begin{APACrefDOI} \doi{10.48550/arXiv.1612.03716} \end{APACrefDOI}
\PrintBackRefs{\CurrentBib}

\bibitem [\protect \citeauthoryear {%
L\BHBI C.~Chen%
\ \protect \BOthers {.}}{%
L\BHBI C.~Chen%
\ \protect \BOthers {.}}{%
{\protect \APACyear {2017}}%
}]{%
chen_rethinking_2017}
\APACinsertmetastar {%
chen_rethinking_2017}%
\begin{APACrefauthors}%
Chen, L\BHBI C.%
, Papandreou, G.%
, Schroff, F.%
\BCBL {}\ \BBA {} Adam, H.%
\end{APACrefauthors}%
\unskip\
\newblock
\APACrefYearMonthDay{2017}{{\APACmonth{12}}}{}.
\newblock
\APACrefbtitle {Rethinking {Atrous} {Convolution} for {Semantic} {Image}
  {Segmentation}.} {Rethinking {Atrous} {Convolution} for {Semantic} {Image}
  {Segmentation}.}
\newblock
\APACaddressPublisher{}{arXiv}.
\newblock
\begin{APACrefURL} [{2023-05-15}]\url{http://arxiv.org/abs/1706.05587}
  \end{APACrefURL}
\newblock
\APACrefnote{arXiv:1706.05587 [cs] version: 3}
\newblock
\begin{APACrefDOI} \doi{10.48550/arXiv.1706.05587} \end{APACrefDOI}
\PrintBackRefs{\CurrentBib}

\bibitem [\protect \citeauthoryear {%
S.~Chen%
\ \BBA {} Biljecki%
}{%
S.~Chen%
\ \BBA {} Biljecki%
}{%
{\protect \APACyear {2023}}%
}]{%
chen_automatic_2023}
\APACinsertmetastar {%
chen_automatic_2023}%
\begin{APACrefauthors}%
Chen, S.%
\BCBT {}\ \BBA {} Biljecki, F.%
\end{APACrefauthors}%
\unskip\
\newblock
\APACrefYearMonthDay{2023}{{\APACmonth{06}}}{}.
\newblock
{\BBOQ}\APACrefatitle {Automatic assessment of public open spaces using street
  view imagery} {Automatic assessment of public open spaces using street view
  imagery}.{\BBCQ}
\newblock
\APACjournalVolNumPages{Cities}{137}{}{104329}.
\newblock
\begin{APACrefURL}
  [{2023-05-15}]\url{https://www.sciencedirect.com/science/article/pii/S0264275123001415}
  \end{APACrefURL}
\newblock
\begin{APACrefDOI} \doi{10.1016/j.cities.2023.104329} \end{APACrefDOI}
\PrintBackRefs{\CurrentBib}

\bibitem [\protect \citeauthoryear {%
Chiang%
\ \protect \BOthers {.}}{%
Chiang%
\ \protect \BOthers {.}}{%
{\protect \APACyear {2023}}%
}]{%
CHIANG2023104679}
\APACinsertmetastar {%
CHIANG2023104679}%
\begin{APACrefauthors}%
Chiang, Y\BHBI C.%
, Liu, H\BHBI H.%
, Li, D.%
\BCBL {}\ \BBA {} Ho, L\BHBI C.%
\end{APACrefauthors}%
\unskip\
\newblock
\APACrefYearMonthDay{2023}{}{}.
\newblock
{\BBOQ}\APACrefatitle {Quantification through deep learning of sky view factor
  and greenery on urban streets during hot and cool seasons} {Quantification
  through deep learning of sky view factor and greenery on urban streets during
  hot and cool seasons}.{\BBCQ}
\newblock
\APACjournalVolNumPages{Landscape and Urban Planning}{232}{}{104679}.
\newblock
\begin{APACrefURL}
  \url{https://www.sciencedirect.com/science/article/pii/S0169204622003280}
  \end{APACrefURL}
\newblock
\begin{APACrefDOI} \doi{https://doi.org/10.1016/j.landurbplan.2022.104679}
  \end{APACrefDOI}
\PrintBackRefs{\CurrentBib}

\bibitem [\protect \citeauthoryear {%
Cigler%
}{%
Cigler%
}{%
{\protect \APACyear {2017}}%
}]{%
cigler_us_2017}
\APACinsertmetastar {%
cigler_us_2017}%
\begin{APACrefauthors}%
Cigler, B\BPBI A.%
\end{APACrefauthors}%
\unskip\
\newblock
\APACrefYearMonthDay{2017}{{\APACmonth{06}}}{}.
\newblock
{\BBOQ}\APACrefatitle {U.{S}. {Floods}: {The} {Necessity} of {Mitigation}}
  {U.{S}. {Floods}: {The} {Necessity} of {Mitigation}}.{\BBCQ}
\newblock
\APACjournalVolNumPages{State and Local Government Review}{49}{2}{127--139}.
\newblock
\begin{APACrefURL} [{2023-05-08}]\url{https://doi.org/10.1177/0160323X17731890}
  \end{APACrefURL}
\newblock
\begin{APACrefDOI} \doi{10.1177/0160323X17731890} \end{APACrefDOI}
\PrintBackRefs{\CurrentBib}

\bibitem [\protect \citeauthoryear {%
Czerniawski%
\ \BBA {} Leite%
}{%
Czerniawski%
\ \BBA {} Leite%
}{%
{\protect \APACyear {2020}}%
}]{%
CZERNIAWSKI2020101131}
\APACinsertmetastar {%
CZERNIAWSKI2020101131}%
\begin{APACrefauthors}%
Czerniawski, T.%
\BCBT {}\ \BBA {} Leite, F.%
\end{APACrefauthors}%
\unskip\
\newblock
\APACrefYearMonthDay{2020}{}{}.
\newblock
{\BBOQ}\APACrefatitle {Automated segmentation of RGB-D images into a
  comprehensive set of building components using deep learning} {Automated
  segmentation of rgb-d images into a comprehensive set of building components
  using deep learning}.{\BBCQ}
\newblock
\APACjournalVolNumPages{Advanced Engineering Informatics}{45}{}{101131}.
\newblock
\begin{APACrefURL}
  \url{https://www.sciencedirect.com/science/article/pii/S1474034620301026}
  \end{APACrefURL}
\newblock
\begin{APACrefDOI} \doi{https://doi.org/10.1016/j.aei.2020.101131}
  \end{APACrefDOI}
\PrintBackRefs{\CurrentBib}

\bibitem [\protect \citeauthoryear {%
Diaz%
\ \protect \BOthers {.}}{%
Diaz%
\ \protect \BOthers {.}}{%
{\protect \APACyear {2022}}%
}]{%
diaz_deriving_2022}
\APACinsertmetastar {%
diaz_deriving_2022}%
\begin{APACrefauthors}%
Diaz, N\BPBI D.%
, Highfield, W\BPBI E.%
, Brody, S\BPBI D.%
\BCBL {}\ \BBA {} Fortenberry, B\BPBI R.%
\end{APACrefauthors}%
\unskip\
\newblock
\APACrefYearMonthDay{2022}{{\APACmonth{04}}}{}.
\newblock
{\BBOQ}\APACrefatitle {Deriving {First} {Floor} {Elevations} within
  {Residential} {Communities} {Located} in {Galveston} {Using} {UAS} {Based}
  {Data}} {Deriving {First} {Floor} {Elevations} within {Residential}
  {Communities} {Located} in {Galveston} {Using} {UAS} {Based} {Data}}.{\BBCQ}
\newblock
\APACjournalVolNumPages{Drones}{6}{4}{81}.
\newblock
\begin{APACrefURL} [{2023-05-08}]\url{https://www.mdpi.com/2504-446X/6/4/81}
  \end{APACrefURL}
\newblock
\begin{APACrefDOI} \doi{10.3390/drones6040081} \end{APACrefDOI}
\PrintBackRefs{\CurrentBib}

\bibitem [\protect \citeauthoryear {%
{FEMA}%
}{%
{FEMA}%
}{%
{\protect \APACyear {2020}}%
}]{%
fema_appendix_2020}
\APACinsertmetastar {%
fema_appendix_2020}%
\begin{APACrefauthors}%
{FEMA}.%
\end{APACrefauthors}%
\unskip\
\newblock
\APACrefYearMonthDay{2020}{{\APACmonth{04}}}{}.
\newblock
{\BBOQ}\APACrefatitle {Appendix {C}: {Lowest} {Floor} {Guide}} {Appendix {C}:
  {Lowest} {Floor} {Guide}}.{\BBCQ}
\newblock
\BIn{} \APACrefbtitle {{NFIP} {Flood} {Insurance} {Manual}} {{NFIP} {Flood}
  {Insurance} {Manual}}\ (\PrintOrdinal{April 2020}\ \BEd).
\newblock
\begin{APACrefURL}
  [{2023-05-10}]\url{https://www.fema.gov/sites/default/files/2020-05/fim_appendix-c-lowest-floor-guide_apr2020.pdf}
  \end{APACrefURL}
\PrintBackRefs{\CurrentBib}

\bibitem [\protect \citeauthoryear {%
FEMA%
}{%
FEMA%
}{%
{\protect \APACyear {2023}}%
}]{%
fima_fima_nodate}
\APACinsertmetastar {%
fima_fima_nodate}%
\begin{APACrefauthors}%
FEMA.%
\end{APACrefauthors}%
\unskip\
\newblock
\APACrefYearMonthDay{2023}{{\APACmonth{05}}}{}.
\newblock
\APACrefbtitle {{FIMA} {NFIP} {Redacted} {Claims} - v1 {\textbar} {FEMA}.gov.}
  {{FIMA} {NFIP} {Redacted} {Claims} - v1 {\textbar} {FEMA}.gov.}
\newblock
\begin{APACrefURL}
  [{2023-05-16}]\url{https://www.fema.gov/openfema-data-page/fima-nfip-redacted-claims-v1}
  \end{APACrefURL}
\PrintBackRefs{\CurrentBib}

\bibitem [\protect \citeauthoryear {%
Feng%
\ \protect \BOthers {.}}{%
Feng%
\ \protect \BOthers {.}}{%
{\protect \APACyear {2022}}%
}]{%
FENG2022101759}
\APACinsertmetastar {%
FENG2022101759}%
\begin{APACrefauthors}%
Feng, Y.%
, Xiao, Q.%
, Brenner, C.%
, Peche, A.%
, Yang, J.%
, Feuerhake, U.%
\BCBL {}\ \BBA {} Sester, M.%
\end{APACrefauthors}%
\unskip\
\newblock
\APACrefYearMonthDay{2022}{}{}.
\newblock
{\BBOQ}\APACrefatitle {Determination of building flood risk maps from LiDAR
  mobile mapping data} {Determination of building flood risk maps from lidar
  mobile mapping data}.{\BBCQ}
\newblock
\APACjournalVolNumPages{Computers, Environment and Urban
  Systems}{93}{}{101759}.
\newblock
\begin{APACrefURL}
  \url{https://www.sciencedirect.com/science/article/pii/S0198971522000035}
  \end{APACrefURL}
\newblock
\begin{APACrefDOI} \doi{https://doi.org/10.1016/j.compenvurbsys.2022.101759}
  \end{APACrefDOI}
\PrintBackRefs{\CurrentBib}

\bibitem [\protect \citeauthoryear {%
G.~Gao%
\ \protect \BOthers {.}}{%
G.~Gao%
\ \protect \BOthers {.}}{%
{\protect \APACyear {2023}}%
}]{%
gao_exploring_2023}
\APACinsertmetastar {%
gao_exploring_2023}%
\begin{APACrefauthors}%
Gao, G.%
, Ye, X.%
, Li, S.%
, Huang, X.%
, Ning, H.%
, Retchless, D.%
\BCBL {}\ \BBA {} Li, Z.%
\end{APACrefauthors}%
\unskip\
\newblock
\APACrefYearMonthDay{2023}{{\APACmonth{05}}}{}.
\newblock
{\BBOQ}\APACrefatitle {Exploring flood mitigation governance by estimating
  first-floor elevation via deep learning and google street view in coastal
  {Texas}} {Exploring flood mitigation governance by estimating first-floor
  elevation via deep learning and google street view in coastal
  {Texas}}.{\BBCQ}
\newblock
\APACjournalVolNumPages{Environment and Planning B: Urban Analytics and City
  Science}{}{}{23998083231175681}.
\newblock
\begin{APACrefURL}
  [{2023-05-17}]\url{https://doi.org/10.1177/23998083231175681}
  \end{APACrefURL}
\newblock
\APACrefnote{Publisher: SAGE Publications Ltd STM}
\newblock
\begin{APACrefDOI} \doi{10.1177/23998083231175681} \end{APACrefDOI}
\PrintBackRefs{\CurrentBib}

\bibitem [\protect \citeauthoryear {%
Y.~Gao%
\ \BBA {} Mosalam%
}{%
Y.~Gao%
\ \BBA {} Mosalam%
}{%
{\protect \APACyear {2018}}%
}]{%
gao_deep_2018}
\APACinsertmetastar {%
gao_deep_2018}%
\begin{APACrefauthors}%
Gao, Y.%
\BCBT {}\ \BBA {} Mosalam, K\BPBI M.%
\end{APACrefauthors}%
\unskip\
\newblock
\APACrefYearMonthDay{2018}{}{}.
\newblock
{\BBOQ}\APACrefatitle {Deep {Transfer} {Learning} for {Image}-{Based}
  {Structural} {Damage} {Recognition}} {Deep {Transfer} {Learning} for
  {Image}-{Based} {Structural} {Damage} {Recognition}}.{\BBCQ}
\newblock
\APACjournalVolNumPages{Computer-Aided Civil and Infrastructure
  Engineering}{33}{9}{748--768}.
\newblock
\begin{APACrefURL}
  [{2022-09-07}]\url{https://onlinelibrary.wiley.com/doi/abs/10.1111/mice.12363}
  \end{APACrefURL}
\newblock
\APACrefnote{\_eprint:
  https://onlinelibrary.wiley.com/doi/pdf/10.1111/mice.12363}
\newblock
\begin{APACrefDOI} \doi{10.1111/mice.12363} \end{APACrefDOI}
\PrintBackRefs{\CurrentBib}

\bibitem [\protect \citeauthoryear {%
Guan%
\ \protect \BOthers {.}}{%
Guan%
\ \protect \BOthers {.}}{%
{\protect \APACyear {2023}}%
}]{%
GUAN2023101975}
\APACinsertmetastar {%
GUAN2023101975}%
\begin{APACrefauthors}%
Guan, F.%
, Fang, Z.%
, Zhang, X.%
, Zhong, H.%
, Zhang, J.%
\BCBL {}\ \BBA {} Huang, H.%
\end{APACrefauthors}%
\unskip\
\newblock
\APACrefYearMonthDay{2023}{}{}.
\newblock
{\BBOQ}\APACrefatitle {Using street-view panoramas to model the decision-making
  complexity of road intersections based on the passing branches during
  navigation} {Using street-view panoramas to model the decision-making
  complexity of road intersections based on the passing branches during
  navigation}.{\BBCQ}
\newblock
\APACjournalVolNumPages{Computers, Environment and Urban
  Systems}{103}{}{101975}.
\newblock
\begin{APACrefURL}
  \url{https://www.sciencedirect.com/science/article/pii/S0198971523000388}
  \end{APACrefURL}
\newblock
\begin{APACrefDOI} \doi{https://doi.org/10.1016/j.compenvurbsys.2023.101975}
  \end{APACrefDOI}
\PrintBackRefs{\CurrentBib}

\bibitem [\protect \citeauthoryear {%
Guo%
\ \protect \BOthers {.}}{%
Guo%
\ \protect \BOthers {.}}{%
{\protect \APACyear {2022}}%
}]{%
guo_large-scale_2022}
\APACinsertmetastar {%
guo_large-scale_2022}%
\begin{APACrefauthors}%
Guo, M.%
, Gong, J.%
\BCBL {}\ \BBA {} Whytlaw, J\BPBI L.%
\end{APACrefauthors}%
\unskip\
\newblock
\APACrefYearMonthDay{2022}{{\APACmonth{02}}}{}.
\newblock
{\BBOQ}\APACrefatitle {Large-scale cloud-based building elevation data
  extraction and flood insurance estimation to support floodplain management}
  {Large-scale cloud-based building elevation data extraction and flood
  insurance estimation to support floodplain management}.{\BBCQ}
\newblock
\APACjournalVolNumPages{International Journal of Disaster Risk
  Reduction}{69}{}{102741}.
\newblock
\begin{APACrefURL}
  [{2023-05-08}]\url{https://www.sciencedirect.com/science/article/pii/S2212420921007020}
  \end{APACrefURL}
\newblock
\begin{APACrefDOI} \doi{10.1016/j.ijdrr.2021.102741} \end{APACrefDOI}
\PrintBackRefs{\CurrentBib}

\bibitem [\protect \citeauthoryear {%
He%
\ \protect \BOthers {.}}{%
He%
\ \protect \BOthers {.}}{%
{\protect \APACyear {2023}}%
}]{%
HE2023104189}
\APACinsertmetastar {%
HE2023104189}%
\begin{APACrefauthors}%
He, J.%
, Zhang, J.%
, Yao, Y.%
\BCBL {}\ \BBA {} Li, X.%
\end{APACrefauthors}%
\unskip\
\newblock
\APACrefYearMonthDay{2023}{}{}.
\newblock
{\BBOQ}\APACrefatitle {Extracting human perceptions from street view images for
  better assessing urban renewal potential} {Extracting human perceptions from
  street view images for better assessing urban renewal potential}.{\BBCQ}
\newblock
\APACjournalVolNumPages{Cities}{134}{}{104189}.
\newblock
\begin{APACrefURL}
  \url{https://www.sciencedirect.com/science/article/pii/S026427512300001X}
  \end{APACrefURL}
\newblock
\begin{APACrefDOI} \doi{https://doi.org/10.1016/j.cities.2023.104189}
  \end{APACrefDOI}
\PrintBackRefs{\CurrentBib}

\bibitem [\protect \citeauthoryear {%
Hosseini%
\ \protect \BOthers {.}}{%
Hosseini%
\ \protect \BOthers {.}}{%
{\protect \APACyear {2023}}%
}]{%
HOSSEINI2023101950}
\APACinsertmetastar {%
HOSSEINI2023101950}%
\begin{APACrefauthors}%
Hosseini, M.%
, Sevtsuk, A.%
, Miranda, F.%
, Cesar, R\BPBI M.%
\BCBL {}\ \BBA {} Silva, C\BPBI T.%
\end{APACrefauthors}%
\unskip\
\newblock
\APACrefYearMonthDay{2023}{}{}.
\newblock
{\BBOQ}\APACrefatitle {Mapping the walk: A scalable computer vision approach
  for generating sidewalk network datasets from aerial imagery} {Mapping the
  walk: A scalable computer vision approach for generating sidewalk network
  datasets from aerial imagery}.{\BBCQ}
\newblock
\APACjournalVolNumPages{Computers, Environment and Urban
  Systems}{101}{}{101950}.
\newblock
\begin{APACrefURL}
  \url{https://www.sciencedirect.com/science/article/pii/S0198971523000133}
  \end{APACrefURL}
\newblock
\begin{APACrefDOI} \doi{https://doi.org/10.1016/j.compenvurbsys.2023.101950}
  \end{APACrefDOI}
\PrintBackRefs{\CurrentBib}

\bibitem [\protect \citeauthoryear {%
Häußler%
}{%
Häußler%
}{%
{\protect \APACyear {2023}}%
}]{%
hausler_gsvpanodepth_nodate}
\APACinsertmetastar {%
hausler_gsvpanodepth_nodate}%
\begin{APACrefauthors}%
Häußler, T.%
\end{APACrefauthors}%
\unskip\
\newblock
\APACrefYearMonthDay{2023}{{\APACmonth{05}}}{}.
\newblock
\APACrefbtitle {{GSVPanoDepth}.} {{GSVPanoDepth}.}
\newblock
\begin{APACrefURL}
  [{2023-05-17}]\url{https://github.com/proog128/GSVPanoDepth.js}
  \end{APACrefURL}
\PrintBackRefs{\CurrentBib}

\bibitem [\protect \citeauthoryear {%
Jain%
\ \protect \BOthers {.}}{%
Jain%
\ \protect \BOthers {.}}{%
{\protect \APACyear {2023}}%
}]{%
Jain_2023_CVPR}
\APACinsertmetastar {%
Jain_2023_CVPR}%
\begin{APACrefauthors}%
Jain, J.%
, Li, J.%
, Chiu, M\BPBI T.%
, Hassani, A.%
, Orlov, N.%
\BCBL {}\ \BBA {} Shi, H.%
\end{APACrefauthors}%
\unskip\
\newblock
\APACrefYearMonthDay{2023}{June}{}.
\newblock
{\BBOQ}\APACrefatitle {OneFormer: One Transformer To Rule Universal Image
  Segmentation} {Oneformer: One transformer to rule universal image
  segmentation}.{\BBCQ}
\newblock
\BIn{} \APACrefbtitle {Proceedings of the IEEE/CVF Conference on Computer
  Vision and Pattern Recognition (CVPR)} {Proceedings of the ieee/cvf
  conference on computer vision and pattern recognition (cvpr)}\
  (\BPG~2989-2998).
\PrintBackRefs{\CurrentBib}

\bibitem [\protect \citeauthoryear {%
Jeon%
\ \BBA {} Woo%
}{%
Jeon%
\ \BBA {} Woo%
}{%
{\protect \APACyear {2023}}%
}]{%
JEON2023104631}
\APACinsertmetastar {%
JEON2023104631}%
\begin{APACrefauthors}%
Jeon, J.%
\BCBT {}\ \BBA {} Woo, A.%
\end{APACrefauthors}%
\unskip\
\newblock
\APACrefYearMonthDay{2023}{}{}.
\newblock
{\BBOQ}\APACrefatitle {Deep learning analysis of street panorama images to
  evaluate the streetscape walkability of neighborhoods for subsidized families
  in Seoul, Korea} {Deep learning analysis of street panorama images to
  evaluate the streetscape walkability of neighborhoods for subsidized families
  in seoul, korea}.{\BBCQ}
\newblock
\APACjournalVolNumPages{Landscape and Urban Planning}{230}{}{104631}.
\newblock
\begin{APACrefURL}
  \url{https://www.sciencedirect.com/science/article/pii/S0169204622002808}
  \end{APACrefURL}
\newblock
\begin{APACrefDOI} \doi{https://doi.org/10.1016/j.landurbplan.2022.104631}
  \end{APACrefDOI}
\PrintBackRefs{\CurrentBib}

\bibitem [\protect \citeauthoryear {%
Juan%
\ \protect \BOthers {.}}{%
Juan%
\ \protect \BOthers {.}}{%
{\protect \APACyear {2020}}%
}]{%
juan_comparing_2020}
\APACinsertmetastar {%
juan_comparing_2020}%
\begin{APACrefauthors}%
Juan, A.%
, Gori, A.%
\BCBL {}\ \BBA {} Sebastian, A.%
\end{APACrefauthors}%
\unskip\
\newblock
\APACrefYearMonthDay{2020}{}{}.
\newblock
{\BBOQ}\APACrefatitle {Comparing floodplain evolution in channelized and
  unchannelized urban watersheds in {Houston}, {Texas}} {Comparing floodplain
  evolution in channelized and unchannelized urban watersheds in {Houston},
  {Texas}}.{\BBCQ}
\newblock
\APACjournalVolNumPages{Journal of Flood Risk Management}{13}{}{e12604}.
\newblock
\begin{APACrefURL}
  \url{https://onlinelibrary.wiley.com/doi/abs/10.1111/jfr3.12604}
  \end{APACrefURL}
\newblock
\begin{APACrefDOI} \doi{10.1111/jfr3.12604} \end{APACrefDOI}
\PrintBackRefs{\CurrentBib}

\bibitem [\protect \citeauthoryear {%
Kaur%
\ \protect \BOthers {.}}{%
Kaur%
\ \protect \BOthers {.}}{%
{\protect \APACyear {2023}}%
}]{%
kaur_large-scale_2023}
\APACinsertmetastar {%
kaur_large-scale_2023}%
\begin{APACrefauthors}%
Kaur, N.%
, Lee, C\BHBI C.%
, Mostafavi, A.%
\BCBL {}\ \BBA {} Mahdavi-Amiri, A.%
\end{APACrefauthors}%
\unskip\
\newblock
\APACrefYearMonthDay{2023}{{\APACmonth{02}}}{}.
\newblock
{\BBOQ}\APACrefatitle {Large-scale building damage assessment using a novel
  hierarchical transformer architecture on satellite images} {Large-scale
  building damage assessment using a novel hierarchical transformer
  architecture on satellite images}.{\BBCQ}
\newblock
\APACjournalVolNumPages{Computer-Aided Civil and Infrastructure
  Engineering}{00}{1-20}{}.
\newblock
\begin{APACrefURL}
  [{2023-05-12}]\url{https://onlinelibrary.wiley.com/doi/abs/10.1111/mice.12981}
  \end{APACrefURL}
\newblock
\APACrefnote{\_eprint:
  https://onlinelibrary.wiley.com/doi/pdf/10.1111/mice.12981}
\newblock
\begin{APACrefDOI} \doi{10.1111/mice.12981} \end{APACrefDOI}
\PrintBackRefs{\CurrentBib}

\bibitem [\protect \citeauthoryear {%
Ki%
\ \BBA {} Lee%
}{%
Ki%
\ \BBA {} Lee%
}{%
{\protect \APACyear {2021}}%
}]{%
KI2021103920}
\APACinsertmetastar {%
KI2021103920}%
\begin{APACrefauthors}%
Ki, D.%
\BCBT {}\ \BBA {} Lee, S.%
\end{APACrefauthors}%
\unskip\
\newblock
\APACrefYearMonthDay{2021}{}{}.
\newblock
{\BBOQ}\APACrefatitle {Analyzing the effects of Green View Index of
  neighborhood streets on walking time using Google Street View and deep
  learning} {Analyzing the effects of green view index of neighborhood streets
  on walking time using google street view and deep learning}.{\BBCQ}
\newblock
\APACjournalVolNumPages{Landscape and Urban Planning}{205}{}{103920}.
\newblock
\begin{APACrefURL}
  \url{https://www.sciencedirect.com/science/article/pii/S0169204620301018}
  \end{APACrefURL}
\newblock
\begin{APACrefDOI} \doi{https://doi.org/10.1016/j.landurbplan.2020.103920}
  \end{APACrefDOI}
\PrintBackRefs{\CurrentBib}

\bibitem [\protect \citeauthoryear {%
Kido%
\ \protect \BOthers {.}}{%
Kido%
\ \protect \BOthers {.}}{%
{\protect \APACyear {2021}}%
}]{%
KIDO2021101281}
\APACinsertmetastar {%
KIDO2021101281}%
\begin{APACrefauthors}%
Kido, D.%
, Fukuda, T.%
\BCBL {}\ \BBA {} Yabuki, N.%
\end{APACrefauthors}%
\unskip\
\newblock
\APACrefYearMonthDay{2021}{}{}.
\newblock
{\BBOQ}\APACrefatitle {Assessing future landscapes using enhanced mixed reality
  with semantic segmentation by deep learning} {Assessing future landscapes
  using enhanced mixed reality with semantic segmentation by deep
  learning}.{\BBCQ}
\newblock
\APACjournalVolNumPages{Advanced Engineering Informatics}{48}{}{101281}.
\newblock
\begin{APACrefURL}
  \url{https://www.sciencedirect.com/science/article/pii/S1474034621000367}
  \end{APACrefURL}
\newblock
\begin{APACrefDOI} \doi{https://doi.org/10.1016/j.aei.2021.101281}
  \end{APACrefDOI}
\PrintBackRefs{\CurrentBib}

\bibitem [\protect \citeauthoryear {%
Lee%
\ \BBA {} Gharaibeh%
}{%
Lee%
\ \BBA {} Gharaibeh%
}{%
{\protect \APACyear {2022}}%
}]{%
lee_assessing_2022}
\APACinsertmetastar {%
lee_assessing_2022}%
\begin{APACrefauthors}%
Lee, C\BHBI C.%
\BCBT {}\ \BBA {} Gharaibeh, N\BPBI G.%
\end{APACrefauthors}%
\unskip\
\newblock
\APACrefYearMonthDay{2022}{{\APACmonth{04}}}{}.
\newblock
{\BBOQ}\APACrefatitle {Assessing surface drainage conditions at the street and
  neighborhood scale: {A} computer vision and flow direction method applied to
  lidar data} {Assessing surface drainage conditions at the street and
  neighborhood scale: {A} computer vision and flow direction method applied to
  lidar data}.{\BBCQ}
\newblock
\APACjournalVolNumPages{Computers, Environment and Urban
  Systems}{93}{}{101755}.
\newblock
\begin{APACrefURL}
  [{2022-01-14}]\url{https://www.sciencedirect.com/science/article/pii/S0198971521001629}
  \end{APACrefURL}
\newblock
\APACrefnote{tex.author+an: 1=highlight}
\newblock
\begin{APACrefDOI} \doi{10.1016/j.compenvurbsys.2021.101755} \end{APACrefDOI}
\PrintBackRefs{\CurrentBib}

\bibitem [\protect \citeauthoryear {%
Letchford%
}{%
Letchford%
}{%
{\protect \APACyear {2023}}%
}]{%
letchford_streetview_2023}
\APACinsertmetastar {%
letchford_streetview_2023}%
\begin{APACrefauthors}%
Letchford, A.%
\end{APACrefauthors}%
\unskip\
\newblock
\APACrefYearMonthDay{2023}{{\APACmonth{05}}}{}.
\newblock
\APACrefbtitle {streetview.} {streetview.}
\newblock
\begin{APACrefURL} [{2023-05-17}]\url{https://github.com/robolyst/streetview}
  \end{APACrefURL}
\newblock
\APACrefnote{original-date: 2016-08-19T01:23:20Z}
\PrintBackRefs{\CurrentBib}

\bibitem [\protect \citeauthoryear {%
Li%
\ \protect \BOthers {.}}{%
Li%
\ \protect \BOthers {.}}{%
{\protect \APACyear {2019}}%
}]{%
li_pedestrian_2019}
\APACinsertmetastar {%
li_pedestrian_2019}%
\begin{APACrefauthors}%
Li, X.%
, Yoshimura, Y.%
, Tu, W.%
\BCBL {}\ \BBA {} Ratti, C.%
\end{APACrefauthors}%
\unskip\
\newblock
\APACrefYearMonthDay{2019}{{\APACmonth{10}}}{}.
\newblock
\APACrefbtitle {A pedestrian level strategy to minimize outdoor sunlight
  exposure in hot summer.} {A pedestrian level strategy to minimize outdoor
  sunlight exposure in hot summer.}
\newblock
\APACaddressPublisher{}{arXiv}.
\newblock
\begin{APACrefURL} [{2023-05-15}]\url{http://arxiv.org/abs/1910.04312}
  \end{APACrefURL}
\newblock
\APACrefnote{arXiv:1910.04312 [physics] version: 1}
\newblock
\begin{APACrefDOI} \doi{10.48550/arXiv.1910.04312} \end{APACrefDOI}
\PrintBackRefs{\CurrentBib}

\bibitem [\protect \citeauthoryear {%
D.~Liu%
\ \protect \BOthers {.}}{%
D.~Liu%
\ \protect \BOthers {.}}{%
{\protect \APACyear {2023}}%
}]{%
LIU2023101924}
\APACinsertmetastar {%
LIU2023101924}%
\begin{APACrefauthors}%
Liu, D.%
, Jiang, Y.%
, Wang, R.%
\BCBL {}\ \BBA {} Lu, Y.%
\end{APACrefauthors}%
\unskip\
\newblock
\APACrefYearMonthDay{2023}{}{}.
\newblock
{\BBOQ}\APACrefatitle {Establishing a citywide street tree inventory with
  street view images and computer vision techniques} {Establishing a citywide
  street tree inventory with street view images and computer vision
  techniques}.{\BBCQ}
\newblock
\APACjournalVolNumPages{Computers, Environment and Urban
  Systems}{100}{}{101924}.
\newblock
\begin{APACrefURL}
  \url{https://www.sciencedirect.com/science/article/pii/S0198971522001685}
  \end{APACrefURL}
\newblock
\begin{APACrefDOI} \doi{https://doi.org/10.1016/j.compenvurbsys.2022.101924}
  \end{APACrefDOI}
\PrintBackRefs{\CurrentBib}

\bibitem [\protect \citeauthoryear {%
L.~Liu%
\ \protect \BOthers {.}}{%
L.~Liu%
\ \protect \BOthers {.}}{%
{\protect \APACyear {2023}}%
}]{%
LIU2023104096}
\APACinsertmetastar {%
LIU2023104096}%
\begin{APACrefauthors}%
Liu, L.%
, Wang, H.%
\BCBL {}\ \BBA {} Duan, J.%
\end{APACrefauthors}%
\unskip\
\newblock
\APACrefYearMonthDay{2023}{}{}.
\newblock
{\BBOQ}\APACrefatitle {How streetscape affects car use: Examining unexamined
  features of urban environment with fine-grained data} {How streetscape
  affects car use: Examining unexamined features of urban environment with
  fine-grained data}.{\BBCQ}
\newblock
\APACjournalVolNumPages{Cities}{132}{}{104096}.
\newblock
\begin{APACrefURL}
  \url{https://www.sciencedirect.com/science/article/pii/S0264275122005352}
  \end{APACrefURL}
\newblock
\begin{APACrefDOI} \doi{https://doi.org/10.1016/j.cities.2022.104096}
  \end{APACrefDOI}
\PrintBackRefs{\CurrentBib}

\bibitem [\protect \citeauthoryear {%
Long%
\ \protect \BOthers {.}}{%
Long%
\ \protect \BOthers {.}}{%
{\protect \APACyear {2015}}%
}]{%
long_fully_2015}
\APACinsertmetastar {%
long_fully_2015}%
\begin{APACrefauthors}%
Long, J.%
, Shelhamer, E.%
\BCBL {}\ \BBA {} Darrell, T.%
\end{APACrefauthors}%
\unskip\
\newblock
\APACrefYearMonthDay{2015}{{\APACmonth{03}}}{}.
\newblock
\APACrefbtitle {Fully {Convolutional} {Networks} for {Semantic}
  {Segmentation}.} {Fully {Convolutional} {Networks} for {Semantic}
  {Segmentation}.}
\newblock
\APACaddressPublisher{}{arXiv}.
\newblock
\begin{APACrefURL} [{2023-05-15}]\url{http://arxiv.org/abs/1411.4038}
  \end{APACrefURL}
\newblock
\APACrefnote{arXiv:1411.4038 [cs]}
\newblock
\begin{APACrefDOI} \doi{10.48550/arXiv.1411.4038} \end{APACrefDOI}
\PrintBackRefs{\CurrentBib}

\bibitem [\protect \citeauthoryear {%
Ma%
\ \protect \BOthers {.}}{%
Ma%
\ \protect \BOthers {.}}{%
{\protect \APACyear {2021}}%
}]{%
MA2021103086}
\APACinsertmetastar {%
MA2021103086}%
\begin{APACrefauthors}%
Ma, X.%
, Ma, C.%
, Wu, C.%
, Xi, Y.%
, Yang, R.%
, Peng, N.%
\BDBL {}Ren, F.%
\end{APACrefauthors}%
\unskip\
\newblock
\APACrefYearMonthDay{2021}{}{}.
\newblock
{\BBOQ}\APACrefatitle {Measuring human perceptions of streetscapes to better
  inform urban renewal: A perspective of scene semantic parsing} {Measuring
  human perceptions of streetscapes to better inform urban renewal: A
  perspective of scene semantic parsing}.{\BBCQ}
\newblock
\APACjournalVolNumPages{Cities}{110}{}{103086}.
\newblock
\begin{APACrefURL}
  \url{https://www.sciencedirect.com/science/article/pii/S0264275120314347}
  \end{APACrefURL}
\newblock
\begin{APACrefDOI} \doi{https://doi.org/10.1016/j.cities.2020.103086}
  \end{APACrefDOI}
\PrintBackRefs{\CurrentBib}

\bibitem [\protect \citeauthoryear {%
Meesuk%
\ \protect \BOthers {.}}{%
Meesuk%
\ \protect \BOthers {.}}{%
{\protect \APACyear {2017}}%
}]{%
MEESUK2017239}
\APACinsertmetastar {%
MEESUK2017239}%
\begin{APACrefauthors}%
Meesuk, V.%
, Vojinovic, Z.%
\BCBL {}\ \BBA {} Mynett, A\BPBI E.%
\end{APACrefauthors}%
\unskip\
\newblock
\APACrefYearMonthDay{2017}{}{}.
\newblock
{\BBOQ}\APACrefatitle {Extracting inundation patterns from flood watermarks
  with remote sensing SfM technique to enhance urban flood simulation: The case
  of Ayutthaya, Thailand} {Extracting inundation patterns from flood watermarks
  with remote sensing sfm technique to enhance urban flood simulation: The case
  of ayutthaya, thailand}.{\BBCQ}
\newblock
\APACjournalVolNumPages{Computers, Environment and Urban
  Systems}{64}{}{239-253}.
\newblock
\begin{APACrefURL}
  \url{https://www.sciencedirect.com/science/article/pii/S019897151630326X}
  \end{APACrefURL}
\newblock
\begin{APACrefDOI} \doi{https://doi.org/10.1016/j.compenvurbsys.2017.03.004}
  \end{APACrefDOI}
\PrintBackRefs{\CurrentBib}

\bibitem [\protect \citeauthoryear {%
Michel-Kerjan%
}{%
Michel-Kerjan%
}{%
{\protect \APACyear {2015}}%
}]{%
michel-kerjan_we_2015}
\APACinsertmetastar {%
michel-kerjan_we_2015}%
\begin{APACrefauthors}%
Michel-Kerjan, E.%
\end{APACrefauthors}%
\unskip\
\newblock
\APACrefYearMonthDay{2015}{{\APACmonth{08}}}{}.
\newblock
{\BBOQ}\APACrefatitle {We must build resilience into our communities} {We must
  build resilience into our communities}.{\BBCQ}
\newblock
\APACjournalVolNumPages{Nature}{524}{7566}{389--389}.
\newblock
\begin{APACrefURL} [{2023-05-08}]\url{https://www.nature.com/articles/524389a}
  \end{APACrefURL}
\newblock
\begin{APACrefDOI} \doi{10.1038/524389a} \end{APACrefDOI}
\PrintBackRefs{\CurrentBib}

\bibitem [\protect \citeauthoryear {%
Middel%
\ \protect \BOthers {.}}{%
Middel%
\ \protect \BOthers {.}}{%
{\protect \APACyear {2019}}%
}]{%
MIDDEL2019122}
\APACinsertmetastar {%
MIDDEL2019122}%
\begin{APACrefauthors}%
Middel, A.%
, Lukasczyk, J.%
, Zakrzewski, S.%
, Arnold, M.%
\BCBL {}\ \BBA {} Maciejewski, R.%
\end{APACrefauthors}%
\unskip\
\newblock
\APACrefYearMonthDay{2019}{}{}.
\newblock
{\BBOQ}\APACrefatitle {Urban form and composition of street canyons: A
  human-centric big data and deep learning approach} {Urban form and
  composition of street canyons: A human-centric big data and deep learning
  approach}.{\BBCQ}
\newblock
\APACjournalVolNumPages{Landscape and Urban Planning}{183}{}{122-132}.
\newblock
\begin{APACrefURL}
  \url{https://www.sciencedirect.com/science/article/pii/S0169204618313550}
  \end{APACrefURL}
\newblock
\begin{APACrefDOI} \doi{https://doi.org/10.1016/j.landurbplan.2018.12.001}
  \end{APACrefDOI}
\PrintBackRefs{\CurrentBib}

\bibitem [\protect \citeauthoryear {%
Minaee%
\ \protect \BOthers {.}}{%
Minaee%
\ \protect \BOthers {.}}{%
{\protect \APACyear {2022}}%
}]{%
9356353}
\APACinsertmetastar {%
9356353}%
\begin{APACrefauthors}%
Minaee, S.%
, Boykov, Y.%
, Porikli, F.%
, Plaza, A.%
, Kehtarnavaz, N.%
\BCBL {}\ \BBA {} Terzopoulos, D.%
\end{APACrefauthors}%
\unskip\
\newblock
\APACrefYearMonthDay{2022}{}{}.
\newblock
{\BBOQ}\APACrefatitle {Image Segmentation Using Deep Learning: A Survey} {Image
  segmentation using deep learning: A survey}.{\BBCQ}
\newblock
\APACjournalVolNumPages{IEEE Transactions on Pattern Analysis and Machine
  Intelligence}{44}{7}{3523-3542}.
\newblock
\begin{APACrefDOI} \doi{10.1109/TPAMI.2021.3059968} \end{APACrefDOI}
\PrintBackRefs{\CurrentBib}

\bibitem [\protect \citeauthoryear {%
Ning%
\ \protect \BOthers {.}}{%
Ning%
\ \protect \BOthers {.}}{%
{\protect \APACyear {2022}}%
}]{%
ning_exploring_2022}
\APACinsertmetastar {%
ning_exploring_2022}%
\begin{APACrefauthors}%
Ning, H.%
, Li, Z.%
, Ye, X.%
, Wang, S.%
, Wang, W.%
\BCBL {}\ \BBA {} Huang, X.%
\end{APACrefauthors}%
\unskip\
\newblock
\APACrefYearMonthDay{2022}{{\APACmonth{07}}}{}.
\newblock
{\BBOQ}\APACrefatitle {Exploring the vertical dimension of street view image
  based on deep learning: a case study on lowest floor elevation estimation}
  {Exploring the vertical dimension of street view image based on deep
  learning: a case study on lowest floor elevation estimation}.{\BBCQ}
\newblock
\APACjournalVolNumPages{International Journal of Geographical Information
  Science}{36}{7}{1317--1342}.
\newblock
\begin{APACrefURL}
  [{2023-04-28}]\url{https://doi.org/10.1080/13658816.2021.1981334}
  \end{APACrefURL}
\newblock
\begin{APACrefDOI} \doi{10.1080/13658816.2021.1981334} \end{APACrefDOI}
\PrintBackRefs{\CurrentBib}

\bibitem [\protect \citeauthoryear {%
{OpenStreetMap contributors}%
}{%
{OpenStreetMap contributors}%
}{%
{\protect \APACyear {2017}}%
}]{%
OpenStreetMap}
\APACinsertmetastar {%
OpenStreetMap}%
\begin{APACrefauthors}%
{OpenStreetMap contributors}.%
\end{APACrefauthors}%
\unskip\
\newblock
\APACrefYearMonthDay{2017}{}{}.
\newblock
\APACrefbtitle {OpenStreetMap.} {Openstreetmap.}
\newblock
\begin{APACrefURL} \url{https://www.openstreetmap.org} \end{APACrefURL}
\PrintBackRefs{\CurrentBib}

\bibitem [\protect \citeauthoryear {%
Pan%
\ \protect \BOthers {.}}{%
Pan%
\ \protect \BOthers {.}}{%
{\protect \APACyear {2022}}%
}]{%
PAN2022101767}
\APACinsertmetastar {%
PAN2022101767}%
\begin{APACrefauthors}%
Pan, C.%
, Wang, J.%
, Chai, W.%
, Kakillioglu, B.%
, {El Masri}, Y.%
, Panagoulia, E.%
\BDBL {}Velipasalar, S.%
\end{APACrefauthors}%
\unskip\
\newblock
\APACrefYearMonthDay{2022}{}{}.
\newblock
{\BBOQ}\APACrefatitle {Capsule network-based semantic segmentation model for
  thermal anomaly identification on building envelopes} {Capsule network-based
  semantic segmentation model for thermal anomaly identification on building
  envelopes}.{\BBCQ}
\newblock
\APACjournalVolNumPages{Advanced Engineering Informatics}{54}{}{101767}.
\newblock
\begin{APACrefURL}
  \url{https://www.sciencedirect.com/science/article/pii/S1474034622002257}
  \end{APACrefURL}
\newblock
\begin{APACrefDOI} \doi{https://doi.org/10.1016/j.aei.2022.101767}
  \end{APACrefDOI}
\PrintBackRefs{\CurrentBib}

\bibitem [\protect \citeauthoryear {%
Perrett%
\ \protect \BOthers {.}}{%
Perrett%
\ \protect \BOthers {.}}{%
{\protect \APACyear {2023}}%
}]{%
PERRETT2023101968}
\APACinsertmetastar {%
PERRETT2023101968}%
\begin{APACrefauthors}%
Perrett, A.%
, Pollard, H.%
, Barnes, C.%
, Schofield, M.%
, Qie, L.%
, Bosilj, P.%
\BCBL {}\ \BBA {} Brown, J\BPBI M.%
\end{APACrefauthors}%
\unskip\
\newblock
\APACrefYearMonthDay{2023}{}{}.
\newblock
{\BBOQ}\APACrefatitle {DeepVerge: Classification of roadside verge biodiversity
  and conservation potential} {Deepverge: Classification of roadside verge
  biodiversity and conservation potential}.{\BBCQ}
\newblock
\APACjournalVolNumPages{Computers, Environment and Urban
  Systems}{102}{}{101968}.
\newblock
\begin{APACrefURL}
  \url{https://www.sciencedirect.com/science/article/pii/S0198971523000315}
  \end{APACrefURL}
\newblock
\begin{APACrefDOI} \doi{https://doi.org/10.1016/j.compenvurbsys.2023.101968}
  \end{APACrefDOI}
\PrintBackRefs{\CurrentBib}

\bibitem [\protect \citeauthoryear {%
Qi%
\ \BBA {} Hankey%
}{%
Qi%
\ \BBA {} Hankey%
}{%
{\protect \APACyear {2021}}%
}]{%
qi_using_2021}
\APACinsertmetastar {%
qi_using_2021}%
\begin{APACrefauthors}%
Qi, M.%
\BCBT {}\ \BBA {} Hankey, S.%
\end{APACrefauthors}%
\unskip\
\newblock
\APACrefYearMonthDay{2021}{{\APACmonth{02}}}{}.
\newblock
{\BBOQ}\APACrefatitle {Using {Street} {View} {Imagery} to {Predict}
  {Street}-{Level} {Particulate} {Air} {Pollution}} {Using {Street} {View}
  {Imagery} to {Predict} {Street}-{Level} {Particulate} {Air}
  {Pollution}}.{\BBCQ}
\newblock
\APACjournalVolNumPages{Environmental Science \&
  Technology}{55}{4}{2695--2704}.
\newblock
\begin{APACrefURL} [{2023-05-15}]\url{https://doi.org/10.1021/acs.est.0c05572}
  \end{APACrefURL}
\newblock
\APACrefnote{Publisher: American Chemical Society}
\newblock
\begin{APACrefDOI} \doi{10.1021/acs.est.0c05572} \end{APACrefDOI}
\PrintBackRefs{\CurrentBib}

\bibitem [\protect \citeauthoryear {%
Sajedi%
\ \BBA {} Liang%
}{%
Sajedi%
\ \BBA {} Liang%
}{%
{\protect \APACyear {2020}}%
}]{%
sajedi_vibration-based_2020}
\APACinsertmetastar {%
sajedi_vibration-based_2020}%
\begin{APACrefauthors}%
Sajedi, S\BPBI O.%
\BCBT {}\ \BBA {} Liang, X.%
\end{APACrefauthors}%
\unskip\
\newblock
\APACrefYearMonthDay{2020}{}{}.
\newblock
{\BBOQ}\APACrefatitle {Vibration-based semantic damage segmentation for
  large-scale structural health monitoring} {Vibration-based semantic damage
  segmentation for large-scale structural health monitoring}.{\BBCQ}
\newblock
\APACjournalVolNumPages{Computer-Aided Civil and Infrastructure
  Engineering}{35}{6}{579--596}.
\newblock
\begin{APACrefURL}
  [{2022-09-07}]\url{https://onlinelibrary.wiley.com/doi/abs/10.1111/mice.12523}
  \end{APACrefURL}
\newblock
\APACrefnote{\_eprint:
  https://onlinelibrary.wiley.com/doi/pdf/10.1111/mice.12523}
\newblock
\begin{APACrefDOI} \doi{10.1111/mice.12523} \end{APACrefDOI}
\PrintBackRefs{\CurrentBib}

\bibitem [\protect \citeauthoryear {%
Strömberg%
}{%
Strömberg%
}{%
{\protect \APACyear {2007}}%
}]{%
stromberg_natural_2007}
\APACinsertmetastar {%
stromberg_natural_2007}%
\begin{APACrefauthors}%
Strömberg, D.%
\end{APACrefauthors}%
\unskip\
\newblock
\APACrefYearMonthDay{2007}{{\APACmonth{09}}}{}.
\newblock
{\BBOQ}\APACrefatitle {Natural {Disasters}, {Economic} {Development}, and
  {Humanitarian} {Aid}} {Natural {Disasters}, {Economic} {Development}, and
  {Humanitarian} {Aid}}.{\BBCQ}
\newblock
\APACjournalVolNumPages{Journal of Economic Perspectives}{21}{3}{199--222}.
\newblock
\begin{APACrefURL}
  [{2023-05-08}]\url{https://www.aeaweb.org/articles?id=10.1257/jep.21.3.199}
  \end{APACrefURL}
\newblock
\begin{APACrefDOI} \doi{10.1257/jep.21.3.199} \end{APACrefDOI}
\PrintBackRefs{\CurrentBib}

\bibitem [\protect \citeauthoryear {%
Taghinezhad%
\ \protect \BOthers {.}}{%
Taghinezhad%
\ \protect \BOthers {.}}{%
{\protect \APACyear {2020}}%
}]{%
taghinezhad_imputation_2020}
\APACinsertmetastar {%
taghinezhad_imputation_2020}%
\begin{APACrefauthors}%
Taghinezhad, A.%
, Friedland, C\BPBI J.%
, Rohli, R\BPBI V.%
\BCBL {}\ \BBA {} Marx, B\BPBI D.%
\end{APACrefauthors}%
\unskip\
\newblock
\APACrefYearMonthDay{2020}{}{}.
\newblock
{\BBOQ}\APACrefatitle {An {Imputation} of {First}-{Floor} {Elevation} {Data}
  for the {Avoided} {Loss} {Analysis} of {Flood}-{Mitigated} {Single}-{Family}
  {Homes} in {Louisiana}, {United} {States}} {An {Imputation} of
  {First}-{Floor} {Elevation} {Data} for the {Avoided} {Loss} {Analysis} of
  {Flood}-{Mitigated} {Single}-{Family} {Homes} in {Louisiana}, {United}
  {States}}.{\BBCQ}
\newblock
\APACjournalVolNumPages{Frontiers in Built Environment}{6}{}{}.
\newblock
\begin{APACrefURL}
  [{2023-05-08}]\url{https://www.frontiersin.org/articles/10.3389/fbuil.2020.00138}
  \end{APACrefURL}
\PrintBackRefs{\CurrentBib}

\bibitem [\protect \citeauthoryear {%
Wu%
\ \protect \BOthers {.}}{%
Wu%
\ \protect \BOthers {.}}{%
{\protect \APACyear {2019}}%
}]{%
WU2019100936}
\APACinsertmetastar {%
WU2019100936}%
\begin{APACrefauthors}%
Wu, H.%
, Yao, L.%
, Xu, Z.%
, Li, Y.%
, Ao, X.%
, Chen, Q.%
\BDBL {}Meng, B.%
\end{APACrefauthors}%
\unskip\
\newblock
\APACrefYearMonthDay{2019}{}{}.
\newblock
{\BBOQ}\APACrefatitle {Road pothole extraction and safety evaluation by
  integration of point cloud and images derived from mobile mapping sensors}
  {Road pothole extraction and safety evaluation by integration of point cloud
  and images derived from mobile mapping sensors}.{\BBCQ}
\newblock
\APACjournalVolNumPages{Advanced Engineering Informatics}{42}{}{100936}.
\newblock
\begin{APACrefURL}
  \url{https://www.sciencedirect.com/science/article/pii/S1474034619302228}
  \end{APACrefURL}
\newblock
\begin{APACrefDOI} \doi{https://doi.org/10.1016/j.aei.2019.100936}
  \end{APACrefDOI}
\PrintBackRefs{\CurrentBib}

\bibitem [\protect \citeauthoryear {%
Xian%
\ \protect \BOthers {.}}{%
Xian%
\ \protect \BOthers {.}}{%
{\protect \APACyear {2017}}%
}]{%
xian_optimal_2017}
\APACinsertmetastar {%
xian_optimal_2017}%
\begin{APACrefauthors}%
Xian, S.%
, Lin, N.%
\BCBL {}\ \BBA {} Kunreuther, H.%
\end{APACrefauthors}%
\unskip\
\newblock
\APACrefYearMonthDay{2017}{{\APACmonth{05}}}{}.
\newblock
{\BBOQ}\APACrefatitle {Optimal house elevation for reducing flood-related
  losses} {Optimal house elevation for reducing flood-related losses}.{\BBCQ}
\newblock
\APACjournalVolNumPages{Journal of Hydrology}{548}{}{63--74}.
\newblock
\begin{APACrefURL}
  [{2023-05-08}]\url{https://www.sciencedirect.com/science/article/pii/S0022169417301373}
  \end{APACrefURL}
\newblock
\begin{APACrefDOI} \doi{10.1016/j.jhydrol.2017.02.057} \end{APACrefDOI}
\PrintBackRefs{\CurrentBib}

\bibitem [\protect \citeauthoryear {%
Zarekarizi%
\ \protect \BOthers {.}}{%
Zarekarizi%
\ \protect \BOthers {.}}{%
{\protect \APACyear {2020}}%
}]{%
zarekarizi_neglecting_2020}
\APACinsertmetastar {%
zarekarizi_neglecting_2020}%
\begin{APACrefauthors}%
Zarekarizi, M.%
, Srikrishnan, V.%
\BCBL {}\ \BBA {} Keller, K.%
\end{APACrefauthors}%
\unskip\
\newblock
\APACrefYearMonthDay{2020}{{\APACmonth{10}}}{}.
\newblock
{\BBOQ}\APACrefatitle {Neglecting uncertainties biases house-elevation
  decisions to manage riverine flood risks} {Neglecting uncertainties biases
  house-elevation decisions to manage riverine flood risks}.{\BBCQ}
\newblock
\APACjournalVolNumPages{Nature Communications}{11}{1}{5361}.
\newblock
\begin{APACrefURL}
  [{2023-05-08}]\url{https://www.nature.com/articles/s41467-020-19188-9}
  \end{APACrefURL}
\newblock
\begin{APACrefDOI} \doi{10.1038/s41467-020-19188-9} \end{APACrefDOI}
\PrintBackRefs{\CurrentBib}

\bibitem [\protect \citeauthoryear {%
Zhang%
\ \protect \BOthers {.}}{%
Zhang%
\ \protect \BOthers {.}}{%
{\protect \APACyear {2020}}%
}]{%
zhang_concrete_2020}
\APACinsertmetastar {%
zhang_concrete_2020}%
\begin{APACrefauthors}%
Zhang, C.%
, Chang, C\BHBI c.%
\BCBL {}\ \BBA {} Jamshidi, M.%
\end{APACrefauthors}%
\unskip\
\newblock
\APACrefYearMonthDay{2020}{}{}.
\newblock
{\BBOQ}\APACrefatitle {Concrete bridge surface damage detection using a
  single-stage detector} {Concrete bridge surface damage detection using a
  single-stage detector}.{\BBCQ}
\newblock
\APACjournalVolNumPages{Computer-Aided Civil and Infrastructure
  Engineering}{35}{4}{389--409}.
\newblock
\begin{APACrefURL}
  [{2022-09-07}]\url{https://onlinelibrary.wiley.com/doi/abs/10.1111/mice.12500}
  \end{APACrefURL}
\newblock
\APACrefnote{\_eprint:
  https://onlinelibrary.wiley.com/doi/pdf/10.1111/mice.12500}
\newblock
\begin{APACrefDOI} \doi{10.1111/mice.12500} \end{APACrefDOI}
\PrintBackRefs{\CurrentBib}

\bibitem [\protect \citeauthoryear {%
Zhao%
\ \protect \BOthers {.}}{%
Zhao%
\ \protect \BOthers {.}}{%
{\protect \APACyear {2023}}%
}]{%
ZHAO2023101915}
\APACinsertmetastar {%
ZHAO2023101915}%
\begin{APACrefauthors}%
Zhao, T.%
, Liang, X.%
, Tu, W.%
, Huang, Z.%
\BCBL {}\ \BBA {} Biljecki, F.%
\end{APACrefauthors}%
\unskip\
\newblock
\APACrefYearMonthDay{2023}{}{}.
\newblock
{\BBOQ}\APACrefatitle {Sensing urban soundscapes from street view imagery}
  {Sensing urban soundscapes from street view imagery}.{\BBCQ}
\newblock
\APACjournalVolNumPages{Computers, Environment and Urban
  Systems}{99}{}{101915}.
\newblock
\begin{APACrefURL}
  \url{https://www.sciencedirect.com/science/article/pii/S0198971522001594}
  \end{APACrefURL}
\newblock
\begin{APACrefDOI} \doi{https://doi.org/10.1016/j.compenvurbsys.2022.101915}
  \end{APACrefDOI}
\PrintBackRefs{\CurrentBib}

\bibitem [\protect \citeauthoryear {%
Zou%
\ \protect \BOthers {.}}{%
Zou%
\ \protect \BOthers {.}}{%
{\protect \APACyear {2021}}%
}]{%
ZOU2021101421}
\APACinsertmetastar {%
ZOU2021101421}%
\begin{APACrefauthors}%
Zou, Z.%
, Zhao, P.%
\BCBL {}\ \BBA {} Zhao, X.%
\end{APACrefauthors}%
\unskip\
\newblock
\APACrefYearMonthDay{2021}{}{}.
\newblock
{\BBOQ}\APACrefatitle {Virtual restoration of the colored paintings on
  weathered beams in the Forbidden City using multiple deep learning
  algorithms} {Virtual restoration of the colored paintings on weathered beams
  in the forbidden city using multiple deep learning algorithms}.{\BBCQ}
\newblock
\APACjournalVolNumPages{Advanced Engineering Informatics}{50}{}{101421}.
\newblock
\begin{APACrefURL}
  \url{https://www.sciencedirect.com/science/article/pii/S1474034621001737}
  \end{APACrefURL}
\newblock
\begin{APACrefDOI} \doi{https://doi.org/10.1016/j.aei.2021.101421}
  \end{APACrefDOI}
\PrintBackRefs{\CurrentBib}

\end{thebibliography}

\end{document}